\documentclass{article}

\usepackage[numbers]{natbib}

\usepackage {arxiv}
\usepackage{cite}
\usepackage{floatrow}
\usepackage{amsmath,amssymb,amsfonts}
\usepackage{caption}
\usepackage{algorithm}
\usepackage{algorithmic}
\usepackage{textcomp}
\usepackage[table]{xcolor}
\usepackage{float}
\usepackage{booktabs}
\usepackage{enumitem}
\usepackage{setspace}
\usepackage[]{graphicx}
\usepackage{amsmath}
\usepackage{amssymb}
\usepackage{epsfig}
\usepackage{epstopdf}

\usepackage{subfigure}
\usepackage{caption}
\usepackage{comment}
\usepackage{multirow}

\title{\LARGE \bf
Deep Imputation of Missing Values in Time Series Health Data: A Review with Benchmarking}

\author{Maksims Kazijevs and Manar D. Samad\\
Department of Computer Science\\
Tennessee State University\\
Nashville, TN, USA\\
\texttt{msamad@tnstate.edu} \\
 }

\begin{document}

\maketitle


\begin{abstract}
The imputation of missing values in multivariate time series (MTS) data \textcolor{black}{is a critical step in ensuring data quality and producing reliable data-driven predictive models. Apart from many statistical approaches, a few recent studies have proposed state-of-the-art deep learning methods to impute missing values in MTS data.} However, the evaluation of these deep methods is limited to one or two data sets, low missing rates, and completely random missing value types. This survey \textcolor{black}{performs six data-centric experiments} to benchmark state-of-the-art deep imputation methods on five time series health data sets. Our extensive analysis reveals that no single imputation method outperforms the others on all five data sets. The imputation performance depends on data types, individual variable statistics, missing value rates, and types. Deep learning methods that jointly perform cross-sectional (across variables) and longitudinal (across time) imputations of missing values in time series data \textcolor{black}{yield statistically better data quality compared to traditional imputation methods. Although computationally expensive, deep learning methods are practical given the current availability of high-performance computing resources, especially when data quality and sample size are highly important in healthcare informatics. Our findings highlight the importance of data-centric selection of imputation methods to optimize data-driven predictive models.} 

\end{abstract}

\keywords {time series, multivariate data, longitudinal imputation, cross-sectional imputation, missing value imputation, deep neural network, electronic health records, sensor data.}

\section{Introduction}
The presence of missing values in real-world data sets is a major obstacle in effectively building machine learning models with reliable predictive outcomes. In the scientific literature, various arbitrary methods are applied to impute the missing values in data only to enable machine learning-based model development. However, the selection of the imputation method regulates the input data quality and ultimately affects the accuracy and robustness of data-driven predictive models. Statisticians and biostatisticians~\citep{vanBuuren2018} have studied the imputation of missing values in tabular data for several decades, followed by recent contributions from advanced machine and deep learning methods~\citep{Biessmann2018, Zhang2018, Sangeetha2020, Madhu2019}.

The field of missing value imputation (MVI) is primarily studied using multivariate tabular data without considering any time-varying variables. Many real-world data sets observed in health sciences and electronic sensor applications commonly include time-varying data collected for multiple variables. In electronic health records (EHR), patient follow-up data are collected over time, where missing values inevitably appear across variables (missing cross-sectional values) and time (missing longitudinal values). In addition to the variable dimensionality, the dimension of time adds an extra layer of computational challenge in estimating missing values. This requires an imputation method to capture the time dependency and variable dependency in missing data. 

Several survey papers have recently been published on time series data imputation methods. Bauer et al. have benchmarked stochastic multiple imputations and deterministic spectral clustering methods for time series data imputation~\citep{Bauer2017}. Shukla et al. have reviewed imputation methods for time series data based on temporal discretization, interpolation, recurrence, attention, and structural invariance~\citep{Shukla2020}. Festag et al. provide a review selectively on generative adversarial networks (GAN) in forecasting and imputing biomedical time series data~\citep{Festag2022}. However, these survey articles do not perform any benchmarking of existing methods.  Sun et al. have reviewed methods for imputing missing values in irregularly sampled health records data~\citep{Sun2020}. They perform mortality prediction to demonstrate the quality of imputed data sets. They compare the imputation accuracy primarily between several basic learning architectures and their variants, including GRU, LSTM, RNN, M-RNN, GRU-D, T-LSTM, and GAN-GRUI. All these imputation methods are proposed in or before 2018. Their work does not include additional experiments on missing data imputations that are required to determine the robustness of imputation algorithms. Khayati et al. evaluate matrix-based and pattern-based methods to impute time series data acquired from sensors~\citep{Khayati2020}. They do not use deep learning methods or perform experiments under different missing value types and rates. Fang et al. have surveyed time series imputation methods based on three deep learning architectures (RNN, GAN, GRU), in addition to other statistical and machine learning-based imputation methods~\citep{Fang2020}. However, this survey does not perform any benchmarking or present experimental results to compare the imputation performance of those deep learning methods. Our paper is one of the first to comprehensively review and benchmark state-of-the-art deep missing value imputation methods on multiple time series health data.

\section {Background review}

The imputation of missing values is conventionally performed by modeling variables or cross-sectional data dependency. However, the cross-sectional imputation is not optimal for time series data because longitudinal data dependency across time is not considered in missing value estimation. For example, electronic health records (EHR) collect patients' follow-up treatments and clinical tests at noncontinuous and uneven time intervals. The heterogeneity in patient samples turns missing values into a stronger function of time than cross-sectional variables. This requirement necessitates methods to optimally estimate missing values in time series data using cross-sectional and longitudinal information. The sections below provide a review of the literature pertaining to cross-sectional and time series data imputation.

\begin{figure}[t]
\centering
\includegraphics[width=0.65\textwidth]  {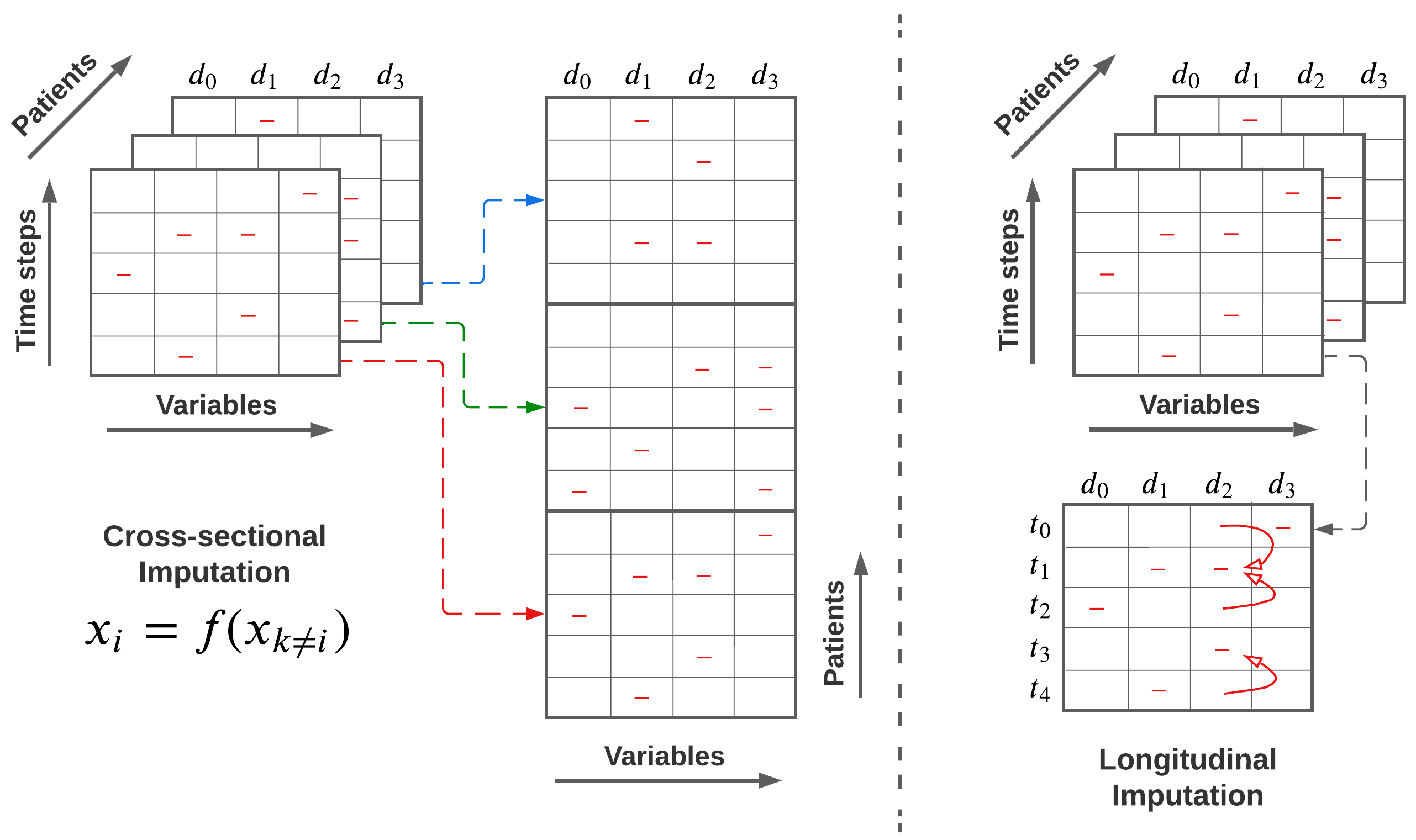}
\caption{Cross-sectional imputation builds regression models based on column variables. Longitudinal imputation leverages time-varying values for individual variables.}
\label{fig:imputation-structure}
\end{figure}

\subsection{Cross-sectional imputation}

A static data set without any time-varying component is the most common data used in the missing value imputation literature. The dependency among variables in columns of a data matrix is often modeled using statistical methods to estimate missing values. This is known as the cross-sectional imputation of missing values. The most common and effective approach for cross-section imputation is based on the statistical method of multiple imputations (MI). The multiple imputations using chained equations (MICE) algorithm~\citep{Resche-Rigon2018, Luo2018} is developed based on the principle of MI~\citep{nassiri2020iterative, vanBuuren2018}. The MICE algorithm generates multiple imputed versions on the same data set, which improves accuracy by reducing uncertainty in data imputation. For the imputation process, MICE builds linear regression models to estimate a dependent variable with missing values as a function of other variables, as shown in Figure~\ref{fig:imputation-structure}. For example, the $j_{th}$ ($x_j$) variable can be estimated from the other (d-1) observed and imputed variables $x_k$ as shown in Equation~\ref{eq:mice}.

\vspace{-10pt}
\begin{eqnarray}
\label{eq:mice}
x_j = f(x_k, W), k \neq j
\end{eqnarray}

Here, W is a vector of trainable regression parameters for the $j_{th}$ variable, where 
\begin{eqnarray}
W = [w_1, w_2, w_3, ..., w_{d-1}]^T
\end{eqnarray}

A recent study has updated the MICE method by replacing linear regressors with several non-linear regressors, including gradient boosting trees (GBT) and multilayer perceptions~\citep{Samad2022}. Results suggest that combining the MI part with ensemble learning (e.g., GBT) techniques provide superior imputation accuracy performance. The MI can capture the variability of different models, while ensemble learning captures variability within a model. 

Apart from statistical learning methods, deep learning has been proposed for the cross-sectional imputation of missing values. One such method trains deep autoencoders on the complete data to encode the values into high-dimensional latent space while the decoder reconstructs the original data from the latent vector to impute missing values~\citep{Choudhury2019}. In recent studies, deep generative methods~\citep{Zhang2018, Camino2019}, convolutional neural networks (CNN)~\citep{Zhuang2019}, and recurrent neural networks (RNN)~\citep{Sangeetha2020} are proposed for missing data imputation. 

\subsubsection {Missing value types} \label{background-mis-types}

The missing value imputation literature on static data sets commonly studies three types of missing values: 1) missing completely at random (MCAR), 2) missing at random (MAR), and 3) missing not at random (MNAR). Three missing value types are statistically defined in the section below. 

We can denote a data matrix, X = \{$X_{obs}$, $X_{miss}$\} that can be composed of  $X_{obs}\subset X$ with the observed values and $X_{mis}\subset X$ with the missing values. Matrix (M = [$m_{ij}$]) represents whether the value at row $i$ and column $j$ is missing or not. Entries when $m_{ij}$ = 0 denote missing values and $m_{ij}$ = 1 denote observed values. A conditional probability distribution of M = 0 (missingness) given X can be defined as $Pr (M = 0 | X_{obs}, X_{mis})$ to model three types of missingness. First, MCAR occurs when the probability of missingness is completely independent of the observed ($X_{obs}$) or missing ($X_{miss}$) data as:

\begin{equation}
    Pr (M=0) = Pr (M=0| X_{obs} , X_{mis}).
\end{equation}

Second, MAR is defined as the missing value type when the probability of missingness depends on the observed data as follows.

\begin{equation}
    Pr (M=0| X_{obs}) = Pr (M=0| X_{obs} , X_{mis}).
\end{equation}

Third, the missing probability depends on the unobserved or missing values themselves in MNAR. 

\begin{equation}
    Pr (M=0| X_{mis}) = Pr (M=0| X_{obs} , X_{mis}).
\end{equation}

However, the time series imputation methods are rarely evaluated against these missing value types. Even for cross-sectional imputation, only a few studies perform a comprehensive analysis of all three missing value types. The MCAR type is chosen as the default type in most studies~\citep{Nikfalazar2020, Gonzalez-Vidal2020, Hegde2019}. In reality, missing values in medical data never appear completely at random because a lab test is missing due to a physician’s recommendation. Physicians do not recommend a lab test when they expect normal results or no significant difference from the previous test within an interval. It has been shown that imputing missing values that are not random is far more challenging and error-prone than imputing MCAR type missing values~\citep{Samad2022}. Therefore, benchmarking missing value imputation methods for time series data requires  a realistic data setting by comparing different missing value types.

\subsection {Time series data imputation}

The imputation of time series data models each variable as a function of time. ECG data can be acquired from one of the multiple electrodes as time-varying multivariate data. Similarly, a clinical variable (e.g., blood pressure) can be observed in the same patient over multiple visits at irregular intervals. The observed values in time can be used to build a predictive model for estimating missing values at other time points, as shown in Figure~\ref{fig:imputation-structure}. 

\subsubsection {Statistical and machine learning based methods}

Statistical methods for imputing missing values in time series data are inspired by the multiple imputation methods developed for cross-sectional imputation. For example, the 3D-MICE method imputes missing values based on the MICE framework by leveraging cross-sectional and longitudinal dependencies in time series data~\citep{Luo2018}. Cross-sectional imputation is performed using standard MICE after flattening time series data. A single-task Gaussian process (GP) is then applied for longitudinal imputation. Both estimates are then combined using a variance-informed weighted average. The 3D-MICE method is primarily tested on one healthcare data set, and it outperforms baseline statistical methods such as mean imputation, MICE, and GP. Recently, Sun et al. have proposed multiple imputations by chained equations with data augmentation (MICE-DA) method to overcome the limitation of separate longitudinal and cross-sectional models of 3D-MICE~\citep{Sun2019}. The MICE-DA method aims to augment flattened cross-sectional data with features extracted from the longitudinal data and then apply the standard MICE method. Xu et al. has proposed the multi-directional multivariate time series (MD-MTS) method for missing value estimation~\citep{Xu2020}. They perform rigorous feature engineering to integrate both temporal and cross-sectional features into a common imputation task. Zhang et al. propose a similar strategy to augment longitudinal features in their xgbooSt MIssing vaLues In timE Series (SMILES) method~\citep{Zhang2020}. Technical details of the three previously mentioned methods (MICE-DA, MD-MTS, and SMILES) are addressed in Section~\ref{DACMI-challenge}. Recently, the time-aware dual-cross-visit (TA-DualCV) method has been proposed to leverage both longitudinal and cross-sectional dependencies within- and cross-patient visits~\citep{Gao2022}. The core part of the method is the dual-cross-visit imputation (DualCV), which captures multivariate and temporal dependencies in cross-visit using chained equations. DualCV consists of two chained equation-based modules: cross-visits feature perspective module (CFP) and cross-visits temporal perspective module (CTP). Both modules utilize Gibbs sampler to impute missing values and are combined subsequently. The time-aware augmentation mechanism then captures patient-specific correlations within each time point by applying the gaussian process (GP) on each patient visit. Lastly, the results from both components, DualCV and GP, are fused using weighted averaging. TA-DualCV is evaluated on three healthcare data sets outperforming time-aware multi-modal autoencoder (TAME)~\citep{Cyin2020}, 3D-MICE, and MICE methods in all experimental settings.

\subsubsection{The DACMI challenge for missing value imputation} \label{DACMI-challenge}

The data analytics challenge on missing data imputation (DACMI)~\citep{Luo2022} has shared de-identified EHR data with training and test partitions for benchmarking time series imputation methods. This challenge has introduced several state-of-the-art statistical and machine learning-based time series data imputation methods, including MD-MTS~\citep{Xu2020}, MICE-DA~\citep{Sun2019}, SMILES~\citep{Zhang2020}, and context-aware time series imputation (CATSI)~\citep{Yin2020}.

In this MD-MTS method, Xu et al. augment longitudinal and cross-sectional features into a new feature set. The feature set includes the following variables: 1) variable values at the current time point, 2) chart time, 3) time stamps, 4) pre and post-values in 3-time stamps, and 5) min, max, and mean values. Then, a tree-based LightGBM regressor for each variable is trained using augmented features to impute missing values. The motivation behind selecting this regressor is that it is less prone to overfitting and more sensitive to outliers. Sun et al. have proposed the MICE-DA method as an improved variant of the 3D-MICE method. This method obtains local temporal features and global patient similarity features for the imputation task. Local temporal features are obtained by taking all pre-and post-variable values within three-time steps. Then, the slope of the change at each time step is estimated using a one-dimensional Gaussian process (1D-GP). Global patient similarity features are obtained by calculating statistical observations from the data set, such as minimum, maximum, percentiles, mean, and standard derivations of individual patient profiles. They combine the local temporal and global patient similarity features in the standard MICE method to impute missing values.  The SMILES method involves three steps: prefilling missing values, feature extraction based on window size, and imputing missing values with the XGBoost model. Local mean and soft impute strategies are used to prefill missing values. However,  selecting which variables will be prefilled with the local mean versus soft impute strategy requires manual work. The XGBoost models are then trained using features extracted based on the window size to impute missing values. Each XGBoost model imputes values for a specific variable, similar to the MICE method. All these newly proposed methods outperform the baseline 3D-MICE method proposed by the organizer of DACMI challenge~\citep{Luo2018}. It is important to note that all the methods \textcolor{black}{submitted to the DACMI challenge} are benchmarked on only one EHR data set, namely the DACMI data set. Therefore, the generalizability of these methods across other time series data sets and data scenarios is unknown.

\subsubsection {Deep time series imputation methods} \label{deep-time-series-imp}

Most of the state-of-the-art time series imputation methods are built upon recurrent neural networks (RNN) and more advanced RNNs such as long short-term memory (LSTM) or gated recurrent unit (GRU) methods. The bidirectional recurrent imputation for time series (BRITS) method is one of the first methods to capture both longitudinal (time) and cross-sectional (variable) dependencies for predicting missing time points~\citep{Cao2018}. BRITS performs imputation using trained LSTM models unfolded in forward and backward directions of time. Thus, bidirectional learning leverages both past and future trends in time series for estimating missing values. Because observations can happen at uneven time intervals, modeling the time gaps in the missing value estimation process is an important contribution of the BRITS method. On the other side, the BRITS method is computationally expensive. BRITS method is evaluated on three data sets related to air quality, health care, and human activities. It is shown that BRITS is superior to RNN-based methods such as GRU-D~\citep{Che2018} and M-RNN~\citep{Yoon2017}. Additionally, BRITS outperforms non-RNN methods, including MICE, spatio-temporal multi-view-based learning (ST-MVL)~\citep{yi2016}, imputeTS~\citep{moritz2017}, matrix factorization (MF), and k-nearest neighbor (KNN) methods. 

Similar to BRITS, the non-autoregressive multiresolution imputation (NAOMI) method uses bidirectional RNN for missing value imputation~\citep{Liu2019}. Furthermore, the authors of NAOMI have introduced the recursive divide and conquer principle. In this principle, two known time steps are identified as pivots, which are used to impute the missing value that falls between them. In the next step, the newly imputed time step is used as the pivot to impute other missing values in between. This process repeats until imputing all missing values. Additionally, the NAOMI method is enhanced with adversarial training. However, NAOMI does not consider time gap information for imputing missing values. The NAOMI method is also evaluated on three data sets: a traffic data set and the other two are trajectory movement data sets. In an additional experiment, the NAOMI method has been tested for robustness on missing rates in data. In this experiment, the authors randomly remove values at a rate ranging from 10\% to 90\% to simulate missing values. NAOMI is compared with several baseline methods, including GRUI~\citep{Luo2018GRU} and MaskGAN~\citep{Fedus2018}. The NAOMI model outperforms these baseline methods in most of the experimental scenarios. Similar GAN-based methods have shown promising results in missing value imputation of time series data~\citep{Zhang2021, Guo2019}.

Yin et al. proposed the CATSI method to capture the global trends of patients. Similar to BRITS architecture, the CATSI method uses bidirectional and cross-sectional dynamics for missing value imputation. To represent the patient health state, they have introduced a "context vector" to learn global temporal dynamics, which improves the missing value imputation accuracy. Also, the CATSI method is aware of time and uses the time gap information similar to the BRITS method. To capture the time gap information between the current time point $s_t$ and the previous time point $s_{t-1}$, the delta matrix $\delta$ is obtained, where $d$ denotes the variable index, $t$ denotes the time index, and $m$ is the mask matrix as shown in Equation~\ref{CATSI-delta}. Equation~\ref{CATSI-delta} shows that if the value at the previous time point is observed ($m^d_{t-1} = 1$), then the time gap at time t ($\delta^d_t$) is the difference between the previous and current time points. When the value at the previous time point is not observed ($m^d_{t-1} = 0$), then the previous time gap value ($\delta^d_{t-1}$) is added to the difference between the previous and current time points to obtain the delta value. 

\vspace{-10pt}
\begin{equation}
\begin{split}
\label{CATSI-delta}
\delta^d_t = \left\{\begin{matrix}
s_t - s_{t-1} + \delta^d_{t-1} & \text{if } t > 0, m^d_{t-1} = 0\\
s_t - s_{t-1} & \text{if } t > 0, m^d_{t-1} = 1\\ 
0 & \text{if } t = 0
\end{matrix}\right.
\end{split}
\end{equation}


If a variable is not observed for a long period of time, the $\delta$ value will be proportionally high. This will produce a low temporal decay coefficient $\gamma_t$, as shown in Equation~\ref{CATSI-gamma}. The data set dependent parameter $\gamma_t$ is obtained via a data-driven estimation method using a trainable parameter $W_{\gamma}$ at the time of training the imputation model.

\vspace{-10pt}
\begin{eqnarray}
\label{CATSI-gamma}
\gamma_t = \text{exp\{-max(}0, W_\gamma \delta_t + b_\gamma)\}
\end{eqnarray}

The decay coefficient $\gamma_t$ is used to fuse the contribution of the last observation of the variable $x^d_{t'}$ and the variable mean $\bar{x}^d$ for initializing the missing value $x^{,d}_t$, as shown in Equation~\ref{CATSI-init}. The missing value will be initialized by the variable mean instead of the last observed value in the remote past. 

\vspace{-10pt}
\begin{eqnarray}
\label{CATSI-init}
x^{,d}_t = \gamma^d_t  x^d_{t'} + (1 - \gamma^d_t) \bar{x}^d
\end{eqnarray}

Another trade-off parameter $\beta_t$ is derived from $\gamma_t$ and the mask matrix and is optimized via a set of trainable parameters ($W_{\beta}, b_{\beta}$), as shown in Equation~\ref{CATSI_beta}.

\vspace{-10pt}
\begin{eqnarray}
\label{CATSI_beta}
\beta_t = \sigma (W_{\beta} [\gamma_t ; m_t] + b_\beta)
\end{eqnarray}
 
 The trade-off parameter $\beta_t$ regulates the contribution of cross-sectional and longitudinal components during the time series imputation model training. For example, the CATSI method uses bidirectional LSTM to obtain longitudinal estimations $\hat{x}$ and MLP for cross-sectional estimations $\hat{z}$ of the missing values. A $\beta_t$ weighted fusion of these estimations yields the final estimate of the missing values $y_t$, as shown in Equation~\ref{CATSI_fusion}. 

\vspace{-10pt}
\begin{eqnarray}
\label{CATSI_fusion}
y_t = \beta_t \odot \hat{z}_t + (1-\beta_t) \odot \hat{x}_t 
\end{eqnarray}

Following the DACMI challenge, several notable imputation methods have emerged, including deep imputer of missing temporal data (DETROIT)~\citep{Yan2019}, Gaussian process variational autoencoder (GP-VAE)~\citep{Fortuin2020}, time-aware multi-modal autoencoder (TAME)~\citep{Cyin2020}. The DETROIT method uses a fully connected neural network with eight hidden layers. To initialize missing values, authors of the DETROIT method manually select the variables for initializing with the local mean or soft impute~\citep{mazumder2010} method. To capture temporal and cross-sectional correlations of variables, the following three data values are incorporated: 1) the values of other variables at the same time step, 2) the values of all variables in the previous and following steps, and 3) the time stamp differences between the four neighboring steps. The DETROIT method is tested only with the DACMI data set, outperforming the baseline 3D-MICE method only. In the GP-VAE method, deep variational autoencoders are used to map time series data with missing values into a latent space. The gaussian process is then used to generate the time series from the low-dimensional latent space. GP-VAE method is tested on two imaging data sets and one tabular medical data set.

Recently, Yin et al. have proposed the TAME method that combines the strengths of a time-aware attention mechanism, bidirectional LSTM architecture, and multi-modal embeddings~\citep{Cyin2020}. First, patient information, such as demographics, diagnosis, medications, and tests, is concatenated and projected into a multi-modal embedding using fully connected layers of neural networks. A bidirectional LSTM architecture is then used to obtain a latent representation of multi-modal embeddings. The time-aware attention mechanism is implemented to capture longitudinal information, which maps value embedding matrices and time gap embedding matrices (the gap between the current time step and the last available time step) into a new space. Finally, missing values are imputed using a fully connected layer, where the inputs are latent vectors from bidirectional LSTM and time-aware attention outcomes. The TAME method takes advantage of time stamp availability and patient information. The method is designed for data sets where patient information, such as demographics, diagnosis, and medications, is available. However, it is also possible to evaluate the TAME method on a data set where such patient information is unavailable, for example, the DACMI data set. However, it needs to be better investigated how this method would perform when extensive patient information is unavailable on multiple benchmark data sets. The TAME method is evaluated on two selective EHR data sets, and it outperforms several recent time series data imputation methods, including 3D-MICE, CATSI, DETROIT, and BRITS.

\subsection {Health records and biomedical sensor data}

\begin{figure}[t]
\centering
\includegraphics[width=0.55\textwidth]  {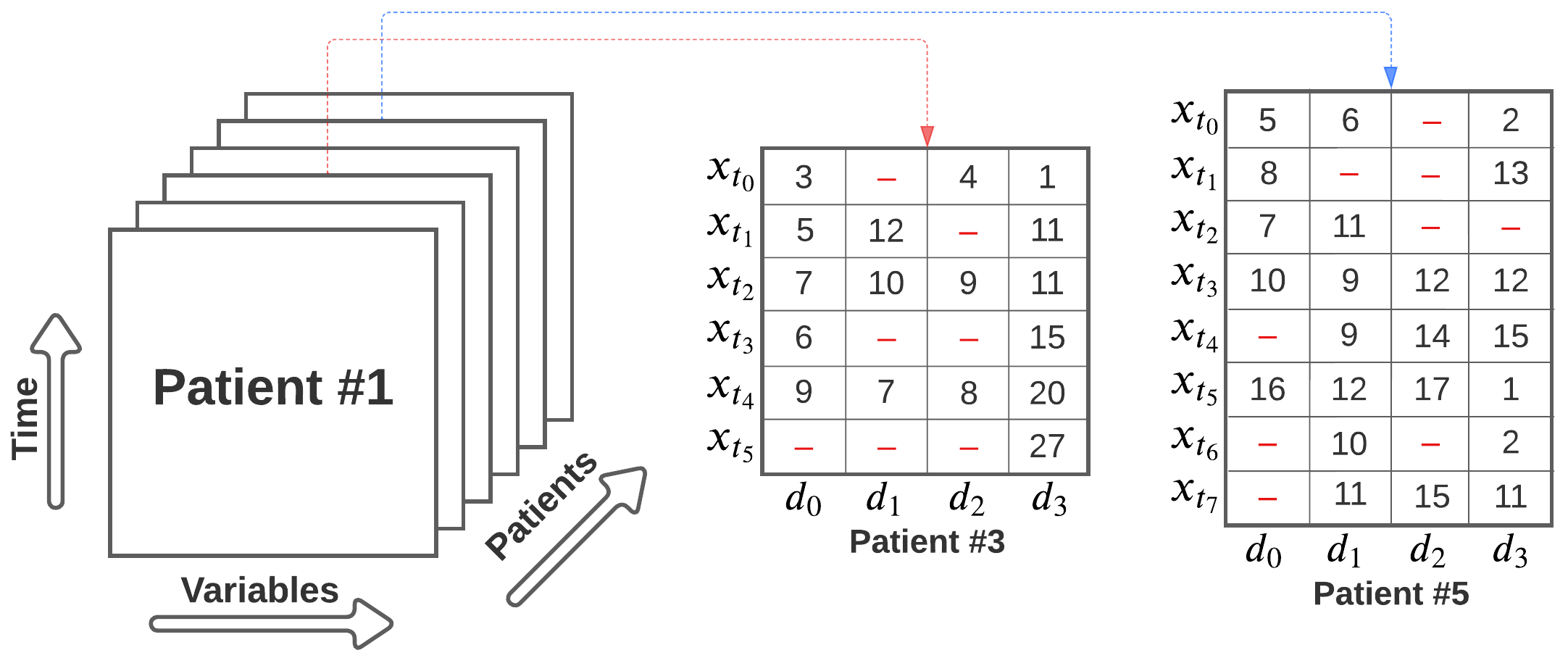}
\caption{Structure of time series health data with three dimensions: patient space, time-space, variable space. Each patient can have a different number of time points or follow-ups, and not all variables are measurements for all time points.}
\label{fig:patient-data}
\end{figure}


Medical centers follow up on patients and store their diagnostic, electronic sensor, and imaging measurements in electronic health records (EHR) for patient care management. A plethora of measurements from hundreds and thousands of patients can be an invaluable resource for retrospective research studies instead of time-consuming clinical trials. However, unlike clinical research trials, where a pre-defined set of parameters are systematically measured and studied over time, EHR data must be curated carefully before using such data in predictive modeling. This is because of the highly complex, incomplete, and unpredictable structure of EHR data that is far from usable in data-driven model development~\citep{Beaulieu-Jones2017, Zhang2018Holistic}. If prepared and processed appropriately, the massive EHR data acquired from a large group of the predominantly sick population may help optimize clinical decision-making and discover novel insights into patient prognosis. Notably, misdiagnosis is the third leading cause of death in the United States~\citep{Makary2016}. Therefore, large EHR data-driven tools and clinical discovery can play an important role in improving the efficacy and precision of medical services. Similarly, biomedical sensor data from electrocardiography (ECG) and electroencephalography (EEG) are widely used in medical practice. 

The application of EHR and biomedical sensor data is receiving growing attention in predictive modeling due to the recent advancement of deep learning in the last decade. For example, Ulloa-Cerna et al. use $\sim$2.2 million ECG signals acquired from $\sim$500,000 patients to predict multiple structural heart conditions using deep learning~\citep{Ulloa2022}. Similarly, novel machine learning methods have enabled health scientists to perform predictive analytics on EHR and medical data to answer unsolved research questions that were previously inconclusive with traditional statistical modeling~\citep{Sherman2016, Jing2016, Samad2018ToF}. The health science literature benefits from numerous applications of machine learning methods, including the estimation of the length of a hospital stay~\citep{Gentimis2017}, real-time mortality prediction~\citep{Nowroozilarki2021}, sepsis prediction~\citep{Scherpf2019, Zhou2021}, blood and arterial pressure estimation ~\citep{Vincent2018, Wang2018, Xing2016}, breathing rate estimation~\citep{Charlton2018}, and more. These research studies can help develop data-driven methods to discover significant clinical risk predictors and make clinical decisions faster and more accurately than those  statistical methods currently used in clinical practice. \textcolor{black}{However, there are three major challenges with utilizing sophisticated machine learning methods to reap the benefits of health record data, as discussed below.}


\subsubsection{Irregular and non-random missing values}
The process of collecting EHR and patient follow-up data is unlike clinical research trials. This is because EHR data are meant for patient care management, not research studies. For example, patient visits and clinical variable measurements can happen irregularly depending on the physician's recommendations and the patient's health status. Therefore, the standard clinical practice collects individual patient data with varying numbers of visits (time points) and measurements taken at varying time intervals based on patient conditions, medical protocols, and administrative reasons~\citep{Luo2018, Sun2019}. See patient \#3 and patient \#5 in Figure~\ref{fig:patient-data}. On the other hand, machine learning methods, by default, expect data samples without irregular time intervals and varying numbers of time points~\citep{Waljee2013, Cheng2020}. In this context, Weerakody et al. present a review of gated recurrent neural networks in several prediction tasks with irregular time series data~\citep{weerakody2021}. Several other reasons can be attributed to missing values in medical data, including when patients fail to visit for a specific test, a staff member fails to record the test result in the database, or the failure of medical sensors or equipment. Likewise, missing values in biomedical sensor data may occur due to broken sensors or loss of data transfer~\citep{Fang2020}. In a study on real-time sensor data acquisition for healthcare monitoring, Michalopoulos et al. mention that the data transmission process cannot be reliable and robust~\citep{Michalopoulos2010}. The reliability and robustness may not be achieved for several reasons, including time synchronization issues, interference, and transmission failures in a wireless sensor network. Therefore, the reason or patterns of missing values are never entirely random. In contrast, the proposed methods for imputing missing values in time series data assume missing values that are missing completely at random. To the best of our knowledge, the effect of missing value types (random versus not at random) has not been studied in the context of time series data imputation. 


\begin{table}[t]
\caption{State-of-the-art methods for imputing missing values in time series data.} 
\scalebox{0.8}{\begin{tabular}{c|c|c|c|c|c|c}
\toprule
  & Method & \begin{tabular}[c]{@{}c@{}}Max. \% of\\ missing data\end{tabular} & \begin{tabular}[c]{@{}c@{}}Number of\\ data sets\end{tabular} & \begin{tabular}[c]{@{}c@{}}Data\\ domain\end{tabular} & \begin{tabular}[c]{@{}c@{}}Irregular\\ time intervals\end{tabular} & \begin{tabular}[c]{@{}c@{}}Downstream\\ classification performance\end{tabular} \\ \midrule 
Luo et al.~\citep{Luo2018} & 3D-MICE & 24\% & One & Healthcare & Yes & No \\ 
Cao et al.~\citep{Cao2018} & BRITS & 13.3\% & Three & Multiple & Yes & Yes \\ 
Liu et al.~\citep{Liu2019} & NAOMI & - & Three & Multiple & No & No \\ 
K. Yin et al.~\citep{Yin2020} & CATSI & 28\% & One & Healthcare & Yes & No \\ 
Yan et al.~\citep{Yan2019} & DETROIT & 7\% & One & Healthcare & Yes & No \\ 
Fortuin et al.~\citep{Fortuin2020} & GP-VAE & 60\% & Three & Multiple & No & Yes \\
C. Yin et al.~\citep{Cyin2020} & TAME & 90\% & Two & Healthcare & Yes & No \\  
Gao et al.~\citep{Gao2022} & TA-DualCV & 90\% & Three & Healthcare & Yes & Yes \\ \bottomrule 
\end{tabular}}
\label{tab:methods-overview}
\end{table}


\subsubsection {Predictive modeling of data with missing values}
Advanced machine learning methods in default settings cannot learn from data with missing values. The selection of an imputation method in the health science literature is always arbitrary without validating or even justifying the selection. In a complete case analysis, the samples with missing entries are excluded to create a data set free of missing values, which is not an optimal way to develop predictive models~\citep{Batista2002}. Such exclusions retain samples from sicker patients who receive more clinical tests than healthier patients. Therefore, such selective population data can yield predictive models with poor generalizability due to patient selection bias~\citep{Weber2017}. The exclusion of data samples or variables can deteriorate the performance of deep learning models that are always data-hungry.

Table~\ref{tab:methods-overview} highlights the state-of-the-art methods proposed for imputing missing values in time series data in terms of the highest missing percentages, the number of data sets used for benchmarking, data domains, irregularity in time series data, and whether downstream classification task is performed to validate the quality of imputed data. Table~\ref{tab:methods-overview} reveals that the state-of-the-art methods lack rigorous benchmarking on multiple data sets and under various experimental scenarios. Benchmarking on only one data set to claim the superiority of a method can be misleading. The only method that performs some rigorous analysis (TA-DualCV) is not a deep learning method. However, none of these recent methods are evaluated for different missing rates and types, considering their viability in real-world settings. Therefore, this paper is inspired by the need to fairly benchmark the deep imputation methods proposed for time series data.

\subsubsection {Missing rates in health data}

A survey of missing value imputation in time series data suggests that the percentage of missing values is assumed constant and low ($\sim$ 10\%). A study on a large EHR data set considers imputing variables up to 50\% missing values~\citep{Beaulieu-Jones2018}. In another study, Samad et al. have found that the most important echocardiographic predictor variable (tricuspid regurgitation maximum velocity - TR max vel) of mortality can be missing for more than 50\% of the patients in a large EHR data set (sample size $>$ 300,000)~\citep{Samad2018}. TR max vel is not commonly measured in patients who receive echocardiographic imaging because its importance is unknown in clinical practice.  

In predictive modeling, variables with more than 50\% missing rates are often excluded based on the assumption that imputing such variables would introduce more impurities and eventually deteriorate the data quality. It is well known that machine learning methods require a large sample size to improve the generalizability of predictive models. The exclusion of samples or variables based on missing rates can deteriorate the overall predictive accuracy of machine learning models.  Therefore, it is imperative to improve the imputation accuracy even at a high missing rate to retain the sample size and discover important predictor variables. The imputation of missing values in Samad et al. is conducted first by linear time interpolation and then using a vanilla MICE (missing value imputation using chained equations) method to learn the variable dependency. To the best of our knowledge, the effect of varying missing value rates on time series data imputation methods is unknown. A robust method is expected to retrain high imputation accuracy even at higher percentages of missing values. 
\section {Methodology}
This section discusses the time series data sets, the selection of missing value imputation methods, and the justification for six experiments conducted to evaluate the state-of-the-art missing value imputation methods for time series data. 


\begin{table}[t]
\caption{Summary of time series health data sets used in the experiments on missing value imputation methods. (.) represents the actual number.}
\scalebox{0.8}{\begin{tabular}{l|c|c|c|c}
\toprule
Data set & Sample size & Num. of variables & Num. of time steps & Domain \\ \midrule
DACMI & 8266 & 13 & Irregular & EHR - ICU patients \\
Sepsis & 2164 & 30 (44) & 20 & Synthetic EHR \\
Hypotension & 3910 & 9 (22) & 48 & Synthetic EHR \\
IEEEPPG & 3096 & 5 & 50 (1000) & PPG, ECG, Acc. Signals \\
Heartbeat & 409 & 61 & 50 (405) & Heart Sound \\ \bottomrule
\end{tabular}}
\label{tab:dataset-overview}
\end{table}


\subsection {Time series health data sets}
We use five health data sets to evaluate the state-of-the-art imputation methods for missing values in time series data. Table~\ref{tab:dataset-overview} summarizes the time series health data used in this paper. The DACMI data set is derived from the MIMIC-III database of electronic health records~\citep{Johnson2016}, and is shared publicly to conduct the DACMI challenge~\citep{Luo2022}. The DACMI data set contains 13 blood laboratory test values of 8267 patients admitted to intensive care units (ICUs). The data set comes with missing value indices and corresponding ground truths. Each patient's lab work is recorded for a specific time point. The DACMI challenge organizer has excluded patients with fewer than ten-time points or follow-up visits. Some lab values have not been measured at the time of the visit, resulting in some natively missing data in the ground-truth data. We exclude those records with natively missing values because our experiments require ground truth for evaluation.

A health gym project publicly shares the acute hypotension and sepsis data sets for developing machine learning models~\citep{Kuo2021}. Patient data are synthetically generated from the MIMIC-III clinical database of ICU patients~\citep{Johnson2016} using generative adversarial neural networks (GANs) \textcolor{black}{to the point that a discriminator model cannot differentiate between synthetic and real patient samples. A three-stage validation is performed to ensure that the two synthetic data sets (sepsis and acute hypotension) sufficiently mimic real patient data~\citep{Kuo2022}. The first stage involves comparing the probability density functions of synthetic and real numeric variables. Binary and categorical variables are compared using side-by-side histogram plots. The second stage performed several statistical tests, including the two-sample Kolmogorov-Smirnov test, two independent Student's t-tests, Snedecor's F-test, and the three-sigma rule test. Using Kendall's rank correlation coefficients, the third stage examines correlations between variables and trends over time.}

\textcolor{black}{The hypotension data set} has 3910 patients measured for 22 variables at one-hour intervals over 48 hours. \textcolor{black}{Nine of these variables are numeric and used in this study.} Non-numeric variables are excluded from our experiments on missing value imputation. The acute hypotension data set does not contain any native missing values. Like the hypotension data set, the sepsis data set is synthetically generated from the MIMIC-III database. It includes information on 2164 patients admitted to intensive care units, with 20 time points recorded for each patient. These time points are derived from 80 hours of data and divided into four-hour segments. At each time point, 44 variables are observed, including 35 numeric, three binary, and six categorical variables. Only 30 numeric variables are selected for our experiments, excluding the binary and categorical variables.

The IEEEPPG data set is shared publicly to conduct the IEEE Signal Processing Cup 2015 for heart rate estimation~\citep{Zhang2014}. The IEEEPPG data set contains the measurements of two-channel photoplethysmographic (PPG) signals obtained by pulse oximeters, three-axis acceleration signals, and one-channel ECG signal, all sampled at 125Hz. The data are recorded from 12 human subjects running at a 15 km/hour peak speed. We use the modified version of the data set obtained from Monash University~\citep{Tan2020} because the original data set is no longer available. The modified version includes five variables: two PPG signals and three-axis acceleration signals only. The signals are segmented into eight seconds windows with six seconds overlap, resulting in 3096 samples with 1000 time steps per sample. We use the first 50 time points for experimental purposes in this paper. 

The heartbeat data set is derived from the PhysioNet and originally shared by the Computing in Cardiology Challenge 2016~\citep{Liu2016, Goldberger2000, Bagnall2018}. The heart sound recordings are collected from four areas of the heart: aortic, pulmonic, tricuspid, and mitral areas. The sound recordings are observed from healthy subjects and pathological patients with confirmed cardiac diagnoses, collected in clinical settings or non-clinical in-home visits. Each recording is truncated to 5 seconds, and a spectrogram of each instance is created with a window size of 0.061 seconds and 70\% overlap, resulting in 405 time steps. The recording consists of 61 dimensions, where each dimension is a frequency band from the spectrogram. The data set includes 409 samples. Each sample is truncated to the first 50 time steps in this paper. 

\subsection {Time series imputation methods}

\textcolor{black}{Our survey identifies recently published methods for multivariate time-series data imputation as of 2022. However, our method selection criteria for benchmarking purposes are as follows:}

\begin{enumerate}
  \item \textcolor{black}{Deep learning-based methods only: We have excluded MD-MTS, MICE-DA, SMILES, TA-DualCV, and 3D-MICE for not involving deep neural networks;}
  \item \textcolor{black}{Availability of source code: Excluded MD-MTS and MICE-DA methods because the source code is unavailable;}
  \item \textcolor{black}{Recent methods only published in the last five years: Excluded GRU-D, M-RNN, and other older methods that more recent methods have consistently outperformed;}
\end{enumerate}

\textcolor{black}{Additionally, we use two baseline methods for benchmarking: 1) the MICE algorithm being one of the most popular imputation methods in health science and 2) LSTM-based imputation as the most basic deep learning method for time series data.} \textcolor{black}{Based on the method selection criteria, we identify eight} state-of-the-art deep imputation methods for time series data. First, we select the BRITS method as one of the first deep learning methods to perform longitudinal and cross-sectional imputation simultaneously. BRITS has proven its performance by outperforming baseline deep learning methods such as GRU-D and M-RNN. Second, we choose the CATSI method because of its novelty in designing a \emph {context vector} that captures a patient's overall health state and the temporal dynamics of the time series data. Moreover, CATSI has proven its superiority by outperforming state-of-the-art imputation methods such as BRITS, 3D-MICE, and MICE. We further split the CATSI method into two parts: CATSI-LSTM and CATSI-MLP, to investigate the contribution of its longitudinal and cross-sectional imputation, respectively. The CATSI-LSTM part includes a bidirectional LSTM structure and the \emph {context vector}. The CATSI-MLP part is a fully connected neural network with two hidden layers for cross-sectional imputation. We also compare the bidirectional CATSI-LSTM method with a unidirectional or conventional LSTM to investigate the benefits of context vectors and bidirectional learning. Third, we include the NAOMI method that utilizes a bidirectional RNN structure similar to the CATSI and BRITS methods. It is one of the first methods that use the divide and conquer principle and adversarial training in the missing value imputation problem. Fourth, we use the recently proposed TAME method that uses the bidirectional LSTM architecture, as discussed in Section~\ref{deep-time-series-imp}. This method outperforms several recently proposed state-of-the-art methods, including BRITS, DETROIT, CATSI, BRNN, and 3D-MICE. The TAME method uses a complete data set with known ground truth data to create the train and test folds. We exclude ground truth for the missing values to avoid information leaks between train and test data splits. \textcolor{black}{Finally, the performances of DETROIT and GP-VAE methods are evaluated in downstream classification tasks and discussed in Section~\ref{exp-classification}.}

\subsection {Experiments}

Our survey of the literature and experience in the DACMI challenge brings forward several findings and research questions. First, a simple piece-wise time interpolation performs better than the sophisticated 3D-MICE method on the DACMI data set~\citep{Samad2019,Luo2022}. It is currently unknown if the claimed superiority of state-of-the-art imputation methods will hold for multiple other data sets. Second, the robustness of time series data imputation methods for varying time lengths, missing rates, and types is currently unknown. The current time series data imputation methods are tested for default low missing rates and data with randomly introduced missing values. Therefore, the robustness and generalizability of these methods across different missing value types and rates remain unknown. Third, it is shown that the best imputation method for a given data set does not consistently perform the best for all variables~\citep{Samad2019}. That is, some variables in time series data are better imputed using cross-sectional imputation methods. This observation needs more rigorous experimentation on the existing methods. Fourth, imputation methods for time series data are evaluated on several imputation accuracy metrics against the known values. However, it is not well known whether superior imputation accuracy translates to superior predictive modeling, such as time series data classification. Model-generated synthetic values for missing entries may alter the variability and predictive information of the data set. We use several strategies to benchmark cross-sectional imputation methods in a recent study~\citep{Samad2022}. This paper performs a fair and rigorous benchmarking of state-of-the-art time series data imputation methods in six experiments using five health-related data sets.      

\subsubsection{Effects of initializing missing values} \label{effect-init} Missing values require some initial values for the deep learning method to start learning the final estimates. 
It is hypothesized that initial values can affect the final convergence of the imputation algorithm. We compare four initialization methods: 1) mean, 2) median, 3) piece-wise time interpolation, and 4) delta initialization considering the time gap between measurements as shown in Equation~\ref{Delta_init}. Here, $\delta$ is the time gap matrix for a patient. We modify the initialization method presented in CATSI and Section~\ref{deep-time-series-imp} to make it independent of the training step while retaining the general idea.  For a long time gap, $\gamma$ is low, and the missing value is approximated by the mean value ($\bar{x}$). Otherwise, when the time gap is shorter, the missing value will be approximated using the value of the next time point, as shown below. 

\begin{equation}
\label{Delta_init}
  \begin{gathered}
    \gamma = \exp^{-\text{max}(0, (\frac{\delta-\mu}{\sigma}))}      \\
    x_t\,_{init} = \gamma * x_{t+1} + (1 - \gamma) * \bar{x}
  \end{gathered}
\end{equation}

4) piece-wise time interpolation, as shown in Equation~\ref{PW_init}, where $s$ is the time stamp, and $z$ is the time point where the value is missing in between $s_t$ and $s_{t+1}$.

\begin{equation}
\label{PW_init}
  \begin{gathered}
    f(z) = x_t + \frac{x_{t+1} - x_t}{s_{t+1} - s_t} * (z - s_t), ~~~z \in [s_t,s_{t+1}]
  \end{gathered}
\end{equation}

\subsubsection{Effects of time series length} It is known that data with longer time series can be challenging to model for recurrent neural networks. This is because the information from the remote past withers away over time during model training. Conversely, a short time length may not have sufficient temporal information for a time series  model. Therefore, we will investigate the effect of time length on the performance of time series data imputation methods. A robust imputation method is expected to show stable and superior performance at shorter or longer time lengths. Each time series is truncated to varying lengths, such as 5, 10, 15, 20, and 25 time steps. If some patients have a time series shorter than any of those lengths, the remainder of the time length is padded with zeros.

\subsubsection{Cross-sectional versus longitudinal versus hybrid imputation methods} \label{effect-cross} Based on our preliminary results in the DACMI challenge, we hypothesize that some variables are more accurately imputed by cross-sectional imputation than longitudinal imputation despite their time-varying trends.  Although integrating cross-sectional and longitudinal imputations as a hybrid method is generally superior for time series data, the superiority can vary from variable to variable. We use piece-wise time interpolation for our experiments as a baseline longitudinal imputation method, as described in Section~\ref{effect-init} and Equation~\ref{PW_init}. We use two cross-sectional imputation methods built on Gradient Boosting Tree (GBT) regressors and chained equations (CE): CE-GBT-I and CE-GBT-II. The CE-GBT-I method trains GBT regressors once for each variable, which are then used to iteratively update missing values in test data by replacing previously imputed values with updates in a chain of regressors. This method is proposed in~\citep{Samad2019}. In contrast, the CE-GBT-II method trains a chain of regressors multiple times by iteratively replacing missing values during the training process. Similar to CE-GBT-I, the CE-GBT-II also iteratively updates missing values.

\begin{algorithm} [t]
\caption{\textcolor{black}{Simulation of three missing value types.}}
\begin{algorithmic}
\STATE Input: $X$: complete data with $n$ samples, $d$ variables ($V$ = $\{x_1, x_2, ..., x_{d}\}$), $t$ time steps,
\\ \qquad  \;\;\; $r$: missing rate, $\tau$: missing type,
\STATE Output: $\chi$: data with missing values
\STATE $\Tilde{X}$ $\leftarrow$ reshape~(X, $n*t$, $d$)
\IF {missing type == missing completely at random}
\STATE [i, j] $\leftarrow$ sample cell indices~($r$)
\STATE $\Tilde{X}$ $\leftarrow$ remove~($\Tilde{X}$, [i, j])
\STATE $\chi$ $\leftarrow$ reshape~($\Tilde{X}$, $n$, $t$, $d$)
\RETURN $\chi$
\ELSE
\STATE low = $r$ / 2, high = 100 - ($r$ / 2)
\IF {missing type == missing at random}
\WHILE{currentMissingRate $<$ $r\%$}
\STATE sample observed variables $|V_{observed}| = 3$ from V
\FOR {q in $V_{observed}$}
\STATE low\_val~(q) $\leftarrow$ Percentile~($\Tilde{X} (:,q)$,~low)
\STATE high\_val~(q) $\leftarrow$ Percentile~($\Tilde{X} (:,q)$,~high)
\STATE $i$ $\leftarrow$ row indices~($<$low\_val(q) $\cup$ $>$high\_val(q))
\ENDFOR
\STATE $V_{missing}$ = $V$ - $V_{osberved}$
\STATE $\Tilde{X}$ $\leftarrow$~remove($V_{missing}$, $i$)
\STATE currentMissingRate $\leftarrow$ $\frac{\textrm{count missing entries} (\Tilde{X})}{n*d} * 100$
\ENDWHILE
\ENDIF
\IF {missing type == missing not at random}
\FOR {p in $V$}
\STATE low\_val (p) $\leftarrow$ Percentile~($\Tilde{X} (:,p)$, low)
\STATE high\_val (p) $\leftarrow$ Percentile~($\Tilde{X} (:,p)$, high)
\STATE values $\leftarrow$ select~($\Tilde{X} (:,p)$,~$<$low\_val $\cup$ $>$high\_val) 
\STATE $\Tilde{X}$ $\leftarrow$ remove~($\Tilde{X} (:,p)$, values)
\ENDFOR
\ENDIF
\STATE $\chi$ $\leftarrow$ reshape~($\Tilde{X}$, $n$, $t$, $d$)
\RETURN $\chi$
\ENDIF
\end{algorithmic}
\end{algorithm}

\subsubsection{Effects of percentage of missing values} \label{exp-percentages} The imputation methods for missing values in time series data generally assume a fixed missing rate between 5\% and 10\%.  For example, The BRITS method is evaluated on three data sets with 10\%, 13\%, and 10\% missing rates. Authors of the DETROIT have used the DACMI data set with 7\% missing values.  We hypothesize that a robust missing value imputation algorithm should also retain its superior performance at higher percentages of missing values. We examine the imputation accuracy on multivariate time series data sets with up to 80\% missing values. 
\subsubsection{Effects of missing value types} \label{exp-types} As mentioned in Section~\ref{background-mis-types}, three missing value types are defined statistically: 1) missing completely at random (MCAR), 2) missing at random (MAR), and 3) missing not at random (MNAR). It is noteworthy that the MCAR type is the most common in the literature because of its simplicity. However, the MCAR type is rarely observed in the real world, with the MNAR being the most practical type. Therefore, the accuracy of existing time series imputation methods adds limited practical value because the robustness of these methods across different missing value types is currently unknown. Our previous work shows that the imputation accuracy of the MNAR type is far worse and more challenging than that reported for the MCAR type~\citep{Samad2022}. Therefore, the robustness of the state-of-the-art imputation methods should be challenged using more complex missing value types. This paper simulates the three missing value types following \textcolor{black}{Algorithm 1~\citep{Samad2022}. The MCAR type is simulated by randomly selecting $r$\% of cell indices and removing values from corresponding positions in time-varying data matrices. The MNAR type is simulated by removing values that are lower than (r/2)-th percentile and higher than [100-(r/2)]-th percentile of a target variable, which accounts for a total of r\% missing rate. The MAR type is achieved iteratively. In each iteration, the same percentiles are obtained on three randomly selected observed variables to identify the indices corresponding to low and high values. These indices are used to remove values in the targeted missing variable set, which does not include observed variables. The iteration continues until achieving a total of r\% missing rate.}

\subsubsection{Classification of imputed data} \label{exp-classification} The imputation of missing values enables machine learning of otherwise inoperable data. \textcolor{black}{However, the quality of model-generated data used to impute missing values} ultimately determines the outcome of machine learning models, such as regression or classification performance. The quality of imputed data is rarely evaluated in downstream classification tasks to confirm if the imputation accuracy translates to classification performance. \textcolor{black}{This experiment compares} the classification accuracy of time series data completed by deep imputation methods ~\textcolor{black}{with that of both ground truth data without missing values and data with missing values without using an imputation method}.

\subsection {Model evaluation} \label{sec-eval}
All data sets are subject to 5\% missing values completely at random (MCAR), except for the experiments related to the effects of the percentage of missing values (Section~\ref{exp-percentages}) and effects of missing value types (Section~\ref{exp-types}). We split each data set into 80\% - 20\%  train and test folds. The train fold with missing values is used to train the models. NAOMI and TAME are the two methods that require ground truth values in the training process to optimize the loss function and model parameters. Other experimental methods do not require ground truth values, as they optimize the model based on the data set's observed values. The trained model imputes missing values on the test data. The imputation accuracy is determined from the model imputed values and the corresponding ground truth on test data. The same train and test data folds are used in all experiments to ensure fair comparison and reproducibility. For evaluation, we obtain normalized root mean square deviation (NRMSD), a widely adopted metric for health data imputation~\citep{Luo2022, Yin2020, Xu2020, Zhang2020, Luo2018}. In NRMSD, the absolute difference between actual ($Y$) and imputed values ($\hat{Y}$) is normalized to bring the variables to the same scales. The result is then multiplied by a mask matrix ($M$) to disregard all non-missing entries, as shown in Equation~\ref{nrmsd}, where $p$ is the patient index, $t$ is the time step index, $d$ is the variable index. 

\vspace{-10pt}
\begin{eqnarray}
\label{nrmsd}
NRMSD(d) = \sqrt{\frac{\sum_{p,t} M_{p,t,d} \left (\frac{ | Y_{p,t,d} - \hat{Y}_{p,t,d}  |}{\textrm{max}(Y_{p,d}) - \textrm{min}(Y_{p,d})} \right)^2 }{\sum_{p,t}M_{p,t,d}}}
\end{eqnarray}

\textcolor{black}{Based on the NRMSD scores, we rank the performance of the imputation or initialization methods for individual variables of a data set. The average rank of a method is obtained by averaging its ranks across all variables. Therefore, the average rank of a method cannot be lower than one or greater than the number of methods.} For our experiment denoted in Section~\ref{exp-classification}, we report the classification performance using a weighted F1 score, as shown in Equation~\ref{f1-score}, where TP, FP, and FN denote true positive, false positive, and false negative predictions, respectively.

\begin{equation}
\label{f1-score}
  \begin{gathered}
    \textrm{F1 score} = \frac{\textrm{TP}}{\textrm{TP} + \frac{1}{2} (\textrm{FP} + \textrm{FN})} \\
  \end{gathered}
\end{equation}
We choose a weighted F1 score for our classification task in Equation~\ref{weighted-f1-score} with the rationale that it is more robust towards an imbalanced data set. 
\begin{equation}
\label{weighted-f1-score}
  \begin{gathered}
    \textrm{Weighted F1 score} = \sum_{i=1}^{N} w_i * \textrm{(F1 score)}_i \textrm{, where} \\ 
    w_i = \frac{\textrm{\# of samples in class } i }{\textrm{Total \# of samples}.}
  \end{gathered}
\end{equation}
Here, $N$ denotes the total number of classes. \textcolor{black}{For statistical comparisons, we perform Wilcoxon signed-rank tests for initialization (Section~\ref{effect-init}) and classification (Section~\ref{exp-classification}) experiments.}

\section{Experimental results}

The sections below detail the results of the six experiments on state-of-the-art deep time series imputation methods. 


\begin{table}[t]
\caption{\textcolor{black}{Effects of missing value initialization methods based on average normalized root mean squared deviation scores (NRMSD). The average NRMSD for an initialization method is obtained across seven imputation models (BRITS, CATSI, CATSI-LSTM, LSTM, CATSI-MLP, NAOMI, MICE) and all variables. The average rank of an initialization method for a data set is obtained across all variables and imputation methods.}}
\resizebox{\textwidth}{!}{\begin{tabular}{c|cc|cc|cc|cc|cc}
\toprule
Dataset & \multicolumn{2}{c|}{DACMI} & \multicolumn{2}{c|}{Sepsis} & \multicolumn{2}{c|}{Hypotension} & \multicolumn{2}{c|}{IEEEPPG} & \multicolumn{2}{c}{Heartbeat} \\ \midrule
Initialization & NRMSD & \begin{tabular}[c]{@{}c@{}}Avg.\\ rank\end{tabular} & NRMSD & \begin{tabular}[c]{@{}c@{}}Avg.\\ rank\end{tabular} & NRMSD & \begin{tabular}[c]{@{}c@{}}Avg.\\ rank\end{tabular} & NRMSD & \begin{tabular}[c]{@{}c@{}}Avg.\\ rank\end{tabular} & NRMSD & \begin{tabular}[c]{@{}c@{}}Avg.\\ rank\end{tabular} \\ \midrule
Delta & 0.383 (0.21) & 2.6 & 0.234 (0.03) & 2.4 & 0.211 (0.06) & 2.1 & 0.281 (0.27) & 2.0 & \textbf{0.204 (0.06)} & \textbf{2.2} \\
Mean & 0.375 (0.19) & 2.9 & \textbf{0.233 (0.03)} & \textbf{1.6} & \textbf{0.211 (0.06)} & \textbf{1.5} & 0.287 (0.27) & 3.2 & 0.205 (0.06) & 2.4 \\
Median & \textbf{0.351 (0.17)} & \textbf{2.1} & 0.234 (0.03) & 2.1 & 0.212 (0.06) & 3.1 & 0.289 (0.27) & 3.8 & 0.205 (0.06) & 2.4 \\
Piece-wise & 0.394 (0.24) & 2.4 & 0.249 (0.05) & 4.0 & 0.215 (0.06) & 3.3 & \textbf{0.219 (0.29)} & \textbf{1.0} & 0.207 (0.07) & 3.0 \\ \bottomrule
\end{tabular}}
\label{tab:misssing-value-init-1}
\end{table}



\begin{table}[t]
\caption{\textcolor{black}{The best initialization methods for an imputation method and data set pair based on normalized root mean squared deviation (NRMSD) scores. The average rank of the best initialization method is obtained by averaging its ranks across all variables in a data set.} PW = Piece-wise interpolation.}
\resizebox{\textwidth}{!}{\begin{tabular}{c|ccc|ccc|ccc|ccc|ccc}
\toprule
Dataset & \multicolumn{3}{c|}{DACMI} & \multicolumn{3}{c|}{Sepsis} & \multicolumn{3}{c|}{Hypotension} & \multicolumn{3}{c|}{IEEEPPG} & \multicolumn{3}{c}{Heartbeat} \\ \midrule
Method & Init. & NRMSD & \begin{tabular}[c]{@{}c@{}}Avg.\\ rank\end{tabular} & Init. & NRMSD & \begin{tabular}[c]{@{}c@{}}Avg.\\ rank\end{tabular} & Init. & NRMSD & \begin{tabular}[c]{@{}c@{}}Avg.\\ rank\end{tabular} & Init. & NRMSD & \begin{tabular}[c]{@{}c@{}}Avg.\\ rank\end{tabular} & Init. & NRMSD & \begin{tabular}[c]{@{}c@{}}Avg.\\ rank\end{tabular} \\ \midrule
BRITS & Delta & 0.266 & 3.6 & \textbf{PW} & \textbf{0.190} & \textbf{1.6} & Mean & 0.163 & 2.1 & Delta & 0.094 & 4.7 & PW & 0.194 & 4.2 \\
CATSI & \textbf{Delta} & \textbf{0.206} & \textbf{1.4} & Delta & 0.202 & 1.9 & \textbf{PW} & \textbf{0.158} & \textbf{1.3} & PW & 0.057 & 2.8 & \textbf{PW} & \textbf{0.139} & \textbf{1.3} \\
CATSI-LSTM & Mean & 0.223 & 2.6 & PW & 0.226 & 3.7 & Mean & 0.173 & 3.0 & \textbf{PW} & \textbf{0.053} & \textbf{1.2} & PW & 0.156 & 2.9 \\
LSTM & PW & 0.241 & 3.6 & Mean & 0.282 & 6.7 & Mean & 0.224 & 5.7 & PW & 0.080 & 4.3 & PW & 0.173 & 3.5 \\
CATSI-MLP & PW & 0.265 & 4.1 & Delta & 0.233 & 4.4 & Mean & 0.204 & 5.0 & PW & 0.294 & 6.0 & Median & 0.164 & 3.4 \\
NAOMI & Median & 0.437 & 6.3 & Median & 0.240 & 4.9 & Delta & 0.218 & 4.9 & PW & 0.054 & 2.0 & Delta & 0.245 & 6.1 \\
\textcolor{black}{MICE} & PW & 0.730 & 6.4 & Median & 0.252 & 4.8 & PW & 0.337 & 6.0 & PW & 0.901 & 7.0 & PW & 0.332 & 6.6 \\ \bottomrule
\end{tabular}}
\label{tab:misssing-value-init-3}
\end{table}



\subsection {Effects of initializing missing values} \label{exp-init}

Table~\ref{tab:misssing-value-init-1} shows the effect of four missing value initialization approaches on five data sets \textcolor{black}{across all imputation methods. Table~\ref{tab:misssing-value-init-3} presents an imputation method-specific selection of the best initialization method and corresponding scores.}

The median initialization achieves the best average rank (2.1) for one (DACMI) out of five data sets with an average NRMSD score of \textcolor{black}{0.351 (0.17)}. Table~\ref{tab:misssing-value-init-3} shows that the NAOMI \textcolor{black}{and MICE methods yield substantially worse imputation accuracy}. Therefore, excluding NAOMI \textcolor{black}{and MICE}, piece-wise initialization ranks as the best initialization method for the DACMI data set with an average NRMSD score of 0.241 \textcolor{black}{(0.02)} and an average rank of 1.2. The exclusion of the NAOMI \textcolor{black}{and MICE methods} turns median initialization into the worst approach, with an average NRMSD score of 0.258 \textcolor{black}{(0.04)} and an average rank of 3.9.

\textcolor{black}{Table~\ref{tab:misssing-value-init-1} shows that} the mean initialization performs the best for the sepsis (NRMSD: \textcolor{black}{0.233 (0.03), average rank: 1.6}) and hypotension (NRMSD: \textcolor{black}{0.211 (0.06), average rank: 1.5}) data sets. These two data sets are synthetically generated at regular time intervals with relatively shorter time series than the other three. The sepsis and hypotension data sets have only 20 and 48 time points for all samples. In contrast to three other data sets, these factors may be important to consider mean initialization over the approaches. 
Table~\ref{tab:misssing-value-init-1} shows that the piece-wise interpolation substantially outperforms all other initialization approaches \textcolor{black}{for the IEEEPPG data set} (NRMSD: \textcolor{black}{0.219 (0.29), average rank: 1.0}). Similarly, the heartbeat data set with a long time series (405-time steps) is truncated to 50-time steps, similar to the IEEEPPG data set. Although delta is the best initialization approach for this data set (NRMSD: \textcolor{black}{0.204 (0.06), average rank: 2.2}), the difference in improvement over other methods (mean, median, and piece-wise) is insignificant. Excluding the NAOMI \textcolor{black}{and MICE} results, piece-wise initialization ranks the best for the heartbeat data set with an average NRMSD of 0.165 \textcolor{black}{(0.02)} and an average rank of 1.8. \textcolor{black}{No initialization method is statistically superior to other methods except for the IEEEPPG data set. The IEEEPPG data set shows that the piece-wise initialization is significantly better than mean, median, and delta initialization (p$<0.05$). We also observe that delta initialization is significantly better than median initialization.} 


\begin{table}[t]
\caption{Effects of time series length on the performance of missing value imputation algorithms. For this experiment, all data sets are subjected to 5\% missing values of the completely-at-random type. The scores represent normalized root squared deviation (NMRSD).}
\scalebox{0.8}{\begin{tabular}{c|c|c|c|c|c|c}
\toprule
Timestep & Method & DACMI & Sepsis & Hypotension & IEEEPPG & Heartbeat \\ \midrule
\multirow{7}{*}{5} & BRITS & 0.844 & \textbf{0.372} & 0.516 & 2.506 & 0.964 \\
 & CATSI & 0.427 & 0.501 & 0.535 & 0.419 & \textbf{0.470} \\
 & CATSI-LSTM & 0.460 & 0.486 & 0.538 & \textbf{0.415} & 0.473 \\
 & LSTM & 0.471 & 0.466 & 0.454 & 0.792 & 0.686 \\
 & CATSI-MLP & \textbf{0.415} & 0.430 & \textbf{0.424} & 0.444 & 0.488 \\
 & NAOMI & 0.493 & 0.483 & 0.465 & 0.440 & 0.516 \\
 & TAME & 0.538 & 0.546 & 0.520 & 0.594 & 0.515 \\ \midrule
\multirow{7}{*}{10} & BRITS & 0.331 & \textbf{0.255} & 0.320 & 0.674 & 0.566 \\
 & CATSI & \textbf{0.267} & 0.280 & 0.364 & 0.210 & \textbf{0.317} \\
 & CATSI-LSTM & 0.296 & 0.308 & 0.354 & \textbf{0.198} & 0.328 \\
 & LSTM & 0.352 & 0.345 & 0.338 & 0.419 & 0.438 \\
 & CATSI-MLP & 0.307 & 0.302 & \textbf{0.299} & 0.351 & 0.362 \\
 & NAOMI & 0.392 & 0.339 & 0.351 & 0.349 & 0.403 \\
 & TAME & 0.346 & 0.335 & 0.318 & 0.318 & 0.340 \\ \midrule
\multirow{7}{*}{15} & BRITS & 0.285 & \textbf{0.213} & \textbf{0.261} & 0.363 & 0.407 \\
 & CATSI & \textbf{0.230} & 0.228 & 0.279 & \textbf{0.126} & \textbf{0.263} \\
 & CATSI-LSTM & 0.258 & 0.252 & 0.264 & 0.132 & 0.285 \\
 & LSTM & 0.325 & 0.305 & 0.294 & 0.438 & 0.344 \\
 & CATSI-MLP & 0.285 & 0.258 & 0.265 & 0.333 & 0.295 \\
 & NAOMI & 0.533 & 0.281 & 0.308 & 0.309 & 0.361 \\
 & TAME & 0.316 & 0.275 & 0.268 & 0.240 & 0.288 \\ \midrule
\multirow{7}{*}{20} & BRITS & 0.270 & \textbf{0.190} & 0.221 & 0.327 & 0.319 \\
 & CATSI & \textbf{0.219} & 0.202 & \textbf{0.218} & \textbf{0.106} & \textbf{0.221} \\
 & CATSI-LSTM & 0.245 & 0.219 & 0.230 & 0.107 & 0.249 \\
 & LSTM & 0.317 & 0.282 & 0.274 & 0.359 & 0.304 \\
 & CATSI-MLP & 0.276 & 0.234 & 0.244 & 0.330 & 0.254 \\
 & NAOMI & 0.693 & 0.245 & 0.284 & 0.293 & 0.332 \\
 & TAME & 0.303 & 0.236 & 0.234 & 0.208 & 0.256 \\ \midrule
\multirow{7}{*}{25} & BRITS & 0.261 & / & 0.201 & 0.198 & 0.286 \\
 & CATSI & \textbf{0.213} & / & \textbf{0.191} & \textbf{0.091} & \textbf{0.194} \\
 & CATSI-LSTM & 0.240 & / & 0.211 & \textbf{0.091} & 0.211 \\
 & LSTM & 0.314 & / & 0.258 & 0.366 & 0.265 \\
 & CATSI-MLP & 0.273 & / & 0.231 & 0.323 & 0.225 \\
 & NAOMI & 0.612 & / & 0.267 & 0.296 & 0.302 \\
 & TAME & 0.299 & / & 0.217 & 0.176 & 0.237 \\ \bottomrule
\end{tabular}}
\label{tab:time-length}
\end{table}

\textcolor{black}{Table~\ref{tab:misssing-value-init-2} in Appendix~A provides comprehensive detail of the performance metrics for all initialization methods for individual imputation models and data sets. This table shows that} the CATSI method with delta initialization has the best imputation performance across all other methods (average rank: \textcolor{black}{3.7}) using the DACMI data set. The sepsis data set shows the best imputation rank (average rank: \textcolor{black}{5.3}) using the BRITS algorithm initialized by the piece-wise time interpolation. The hypotension data set is the best imputed by the CATSI algorithm, initialized by the piece-wise time interpolation (average rank: \textcolor{black}{4.1}). The same algorithm and initialization combination ranks the best for the heartbeat data set (average rank: \textcolor{black}{3.9}). However, the LSTM part of the CATSI algorithm (CATSI-LSTM) with piece-wise time interpolation yields the best imputation accuracy for the IEEEPPG data set (average rank: 1.2). In other words, the missing values of the IEEEPPG data set are better predicted across time than across the variables. This observation is supported by the finding that the IEEEPPG also has the lowest correlations between variables. Overall, the CATSI method with piece-wise interpolation ranks the best (rank: 1), considering its imputation performance across all data sets. \textcolor{black}{At the same time, non-deep-learning MICE method ranks the worst for any initialization settings.} These rankings suggest that advanced imputation methods take lesser benefits from any superior approach for initializing missing values. 

\subsection {Effects of time series length} 
Table~\ref{tab:time-length} compares the imputation accuracy for varying time-series lengths. Mean initialization is used, except for the TAME method, which does not explicitly initialize missing values. To achieve a constant time length for fair comparisons, longer and shorter time series data are truncated and padded with zeros, respectively. The cross-sectional imputation method (CATSI-MLP) is the best for short time series data (five time points) on the DACMI (NRMSD: 0.415) and hypotension (NRMSD: 0.424) data sets. The BRITS method performs the best for the sepsis data set for all time series lengths.
The CATSI method performs the best with the heartbeat data set for all time series lengths and is superior on the IEEEPPG data set when the time series length is 15 (NRMSD: 0.126) and 20 (NRMSD: 0.106) time points. When the time series has five and ten time points, the CATSI-LSTM method shows the best imputation accuracy with NRMSD scores of 0.415 and 0.198, respectively. CATSI and CATSI-LSTM rank the best for 25 time points (NRMSD: 0.091). Similar results are observed on the DACMI data set, where CATSI outperforms other methods when the time series length is ten or more. For the hypotension data set, the results are mixed. When the time series is short (five and ten time points), the cross-sectional CATSI-MLP imputation method outperforms other methods. When the length of the time series is 15, BRITS performs the best (NRMSD: 0.261). However, the difference between the BRITS, CATSI-LSTM, CATSI-MLP, and TAME methods is insignificant. The CATSI method ranks the best when the hypotension data set has 20 (NRMSD: 0.218) and 25 (NRMSD: 0.191) time steps. Overall, the CATSI appears to be the best method when the time series has more than 15 time points.


\subsection {Effect of cross-sectional versus longitudinal versus hybrid imputation methods} \label{exp-cross}

The sections below present data set-specific findings about the performance of different imputation strategies.


\begin{table}[t]
\caption{Variable specific imputation error for the DACMI data set based on normalized root mean square deviation scores. Mean initialization is used for all methods except TAME and piece-wise time interpolation.} 
\resizebox{\textwidth}{!}{\begin{tabular}{c|c|c|c|c|c|c|c|c|c|c|c|c|c|c|c}
\toprule
Method & PCL & PK & PLCO2 & PNA & HCT & HGB & MCV & PLT & WBC & RDW & PBUN & PCRE & PGLU & \begin{tabular}[c]{@{}c@{}}Avg.\\ NRMSD\end{tabular} & \begin{tabular}[c]{@{}c@{}}Avg.\\ rank\end{tabular} \\ \midrule
BRITS & 0.191 & 0.269 & \textbf{0.219} & \textbf{0.202} & 0.145 & 0.146 & 0.395 & 0.351 & 0.269 & 0.431 & 0.264 & 0.293 & 0.293 & 0.267 & 3.77 \\
CATSI & \textbf{0.187} & 0.261 & 0.226 & 0.206 & \textbf{0.137} & \textbf{0.137} & \textbf{0.251} & \textbf{0.176} & 0.219 & 0.222 & 0.183 & \textbf{0.220} & 0.265 & \textbf{0.207} & \textbf{1.62} \\
CATSI-LSTM & 0.210 & \textbf{0.260} & 0.231 & 0.233 & 0.218 & 0.215 & 0.254 & 0.177 & 0.219 & \textbf{0.216} & 0.184 & 0.221 & \textbf{0.263} & 0.223 & 2.85 \\
LSTM & 0.308 & 0.299 & 0.304 & 0.301 & 0.285 & 0.286 & 0.319 & 0.322 & 0.299 & 0.339 & 0.323 & 0.344 & 0.299 & 0.310 & 6.92 \\
CATSI-MLP & 0.227 & 0.297 & 0.279 & 0.236 & 0.142 & 0.147 & 0.304 & 0.308 & 0.302 & 0.334 & 0.307 & 0.297 & 0.306 & 0.268 & 4.69 \\
NAOMI & 0.940 & 0.428 & 0.378 & 1.114 & 0.453 & 0.504 & 1.269 & 0.343 & 0.328 & 1.222 & 0.353 & 0.354 & 0.334 & 0.617 & 8.77 \\
TAME & 0.262 & 0.319 & 0.291 & 0.267 & 0.224 & 0.228 & 0.332 & 0.278 & 0.311 & 0.338 & 0.300 & 0.331 & 0.339 & 0.294 & 6.46 \\
CE-GBT-I & 0.246 & 0.334 & 0.310 & 0.247 & 0.150 & 0.148 & 1.370 & 0.962 & 0.858 & 1.546 & 0.647 & 0.931 & 0.442 & 0.630 & 8.38 \\
CE-GBT-II & 0.244 & 0.334 & 0.308 & 0.245 & 0.149 & 0.149 & 1.369 & 0.970 & 0.836 & 1.539 & 0.631 & 0.919 & 0.421 & 0.624 & 7.85 \\
Piece-wise & 0.222 & 0.272 & 0.225 & 0.234 & 0.233 & 0.232 & 0.258 & 0.181 & \textbf{0.214} & 0.224 & \textbf{0.174} & 0.229 & 0.295 & 0.230 & 3.69 \\ \bottomrule
\end{tabular}}
\label{tab:cross-time-dacmi}
\end{table}

\subsubsection{DACMI data set}

Table~\ref{tab:cross-time-dacmi} reveals that most variables (eight out of 13) in the DACMI data set are best imputed by a hybrid imputation method (BRITS or CATSI) that combines cross-sectional and longitudinal learning. The remaining five variables (PK, WBC,  RDW, PBUN, and PGLU) are best imputed using longitudinal methods (CATSI-LSTM or piece-wise). However, two variables (HCT and HGB) correlate more with other variables than their time-series trend. A pair-wise cross-correlation analysis shows a high correlation (0.96) between  HCT and HGB. The cross-sectional CATSI-MLP-based imputation method performs better than longitudinal imputation methods for these two variables. \textcolor{black}{Figures~\ref{fig:dac}(a)-(b) in Appendix B shows that higher correlation scores between variables yield better imputation accuracy (lower NRMSD) using the cross-sectional MLP method. These figures show that the imputation accuracy (NRMSD scores) is almost constant across varying correlation values when using a longitudinal imputation method (CATSI-LSTM).} Overall, the cross-sectional imputation method ranks second and third for HCT and HGB variables, respectively. Similarly, the correlation between PCL and PNA time-series trends is moderate (0.70), which may not be strong enough to give the cross-sectional imputation method an edge over its longitudinal imputation counterpart. All other variable pairs have a weak correlation (\textless 0.6), which explains the suboptimal performance of cross-sectional imputation methods for these variables. Two variables (WBC and PBUN) show the best imputation accuracy with the piece-wise time interpolation method, suggesting that these variables are more  predictable from their time-series trends. The cross-sectional imputation method performs poorly for these two variables, possibly because these variables are poorly correlated with other variables.  Several other variables (PK, RDW, and PGLU) are best imputed using the longitudinal part of the hybrid CATSI method (CATSI-LSTM). Therefore, the pairwise correlation between variables may provide an important guideline in selecting an appropriate imputation method.

\subsubsection{Sepsis data set}

The sepsis data set indicates that hybrid imputation methods are the best on 17 out of 30 variables. One variable (PaO2) shows the best imputation accuracy with the longitudinal CATSI-LSTM method (NRMSD: 0.199). The remaining 12 variables are best imputed with cross-sectional imputation methods despite being part of a time series data set. For example, the SysBP and meanBP variables have a 0.68 correlation, and SysBP is best imputed using the CE-GBT-II method. Intuitively, mean, systolic, and diastolic blood pressure variables can be correlated. The MeanBP and SysBP variables with a 0.76 correlation are best imputed using the CE-GBT-I and CE-GBT-II cross-sectional imputation methods, respectively. The CE-GBT-I method also performs best for several variables, including Na, Cl, CO2, BE, HCO3, SGPT, and PaCO2. The remaining variables, such as DiaBP, Creatinine, and SGOT, are best imputed by the CE-GBT-II method. Some variables that are best imputed by cross-sectional methods have a moderate correlation with each other: Na-Cl (0.65), CO2-BE (0.64), CO2-HCO3 (0.63), and SGOT-SGPT (0.79). It is important to note that the sepsis data set has a relatively short time series with 20 time steps. This may contribute to the superior performance of cross-sectional imputation methods compared to their hybrid and longitudinal counterparts.


\begin{table}[t]
\caption{Variable specific imputation error for the hypotension data set based on normalized root mean square deviation scores. Mean initialization is used for all methods except TAME and piece-wise time interpolation.} 
\scalebox{0.8}{\begin{tabular}{c|c|c|c|c|c|c|c|c|c|c|c}
\toprule
Method & MAP & DBP & SBP & Urine & ALT & AST & PO2 & Lactic & Serum & \begin{tabular}[c]{@{}c@{}}Avg.\\ NRMSD\end{tabular} & \begin{tabular}[c]{@{}c@{}}Avg.\\ rank\end{tabular} \\ \midrule
BRITS & 0.122 & \textbf{0.129} & \textbf{0.161} & 0.175 & 0.173 & 0.169 & \textbf{0.170} & \textbf{0.179} & 0.185 & 0.163 & 2.22 \\
CATSI & 0.124 & 0.141 & \textbf{0.161} & \textbf{0.161} & \textbf{0.162} & \textbf{0.163} & 0.171 & 0.180 & \textbf{0.160} & \textbf{0.158} & \textbf{1.89} \\
CATSI-LSTM & 0.182 & 0.179 & 0.181 & 0.165 & 0.166 & 0.166 & 0.172 & 0.183 & 0.163 & 0.173 & 3.56 \\
LSTM & 0.223 & 0.222 & 0.226 & 0.219 & 0.222 & 0.223 & 0.213 & 0.228 & 0.239 & 0.224 & 7.11 \\
CATSI-MLP & 0.133 & 0.157 & 0.190 & 0.211 & 0.216 & 0.231 & 0.226 & 0.229 & 0.240 & 0.204 & 6.67 \\
NAOMI & 0.226 & 0.221 & 0.223 & 0.207 & 0.213 & 0.213 & 0.212 & 0.227 & 0.222 & 0.218 & 6.11 \\
TAME & 0.143 & 0.150 & 0.171 & 0.190 & 0.197 & 0.181 & 0.213 & 0.207 & 0.195 & 0.183 & 4.44 \\
CE-GBT-I & \textbf{0.119} & 0.137 & 0.199 & 0.256 & 0.268 & 0.317 & 0.222 & 0.270 & 0.774 & 0.285 & 7.28 \\
CE-GBT-II & \textbf{0.119} & 0.137 & 0.199 & 0.256 & 0.268 & 0.317 & 0.222 & 0.270 & 0.774 & 0.285 & 7.28 \\
Piece-wise & 0.249 & 0.246 & 0.235 & 0.252 & 0.243 & 0.225 & 0.256 & 0.241 & 0.201 & 0.239 & 8.44 \\ \bottomrule
\end{tabular}}
\label{tab:cross-time-hypotension}
\end{table}


\begin{table}[t!]
\caption{Variable specific imputation error for the IEEEPPG data set based on normalized root mean square deviation scores. Mean initialization is used for all methods except TAME and piece-wise time interpolation.}
\scalebox{0.8}{\begin{tabular}{c|c|c|c|c|c|c|c}
\toprule
Method & PPG1 & PPG2 & Acc. X & Acc. Y & Acc. Z & \begin{tabular}[c]{@{}c@{}}Avg.\\ NRMSD\end{tabular} & \begin{tabular}[c]{@{}c@{}}Avg.\\ rank\end{tabular} \\ \midrule
BRITS & 0.067 & 0.071 & 0.091 & 0.104 & 0.136 & 0.094 & 4.00 \\
CATSI & 0.067 & 0.060 & 0.065 & 0.065 & 0.068 & 0.065 & 2.00 \\
CATSI-LSTM & 0.069 & 0.064 & 0.067 & 0.066 & 0.072 & 0.068 & 3.20 \\
LSTM & 0.310 & 0.305 & 0.335 & 0.304 & 0.333 & 0.317 & 8.00 \\
CATSI-MLP & 0.297 & 0.292 & 0.293 & 0.298 & 0.293 & 0.295 & 7.00 \\
NAOMI & 0.282 & 0.283 & 0.264 & 0.269 & 0.267 & 0.273 & 6.00 \\
TAME & 0.137 & 0.130 & 0.137 & 0.134 & 0.134 & 0.134 & 4.80 \\
CE-GBT-I & 0.398 & 0.417 & 0.825 & 1.300 & 0.992 & 0.787 & 9.40 \\
CE-GBT-II & 0.396 & 0.426 & 0.829 & 1.338 & 0.945 & 0.787 & 9.60 \\
Piece-wise & \textbf{0.059} & \textbf{0.052} & \textbf{0.050} & \textbf{0.050} & \textbf{0.057} & \textbf{0.054} & \textbf{1.00} \\ \bottomrule
\end{tabular}}
\label{tab:cross-time-ieeeppg}
\end{table}


\begin{figure*}[t!]
\centering
\vspace{10pt}
\subfigure[DACMI] { \includegraphics[trim=0.23cm 0.2cm 0.24cm 0.25cm, clip,
width=0.315\textwidth]  {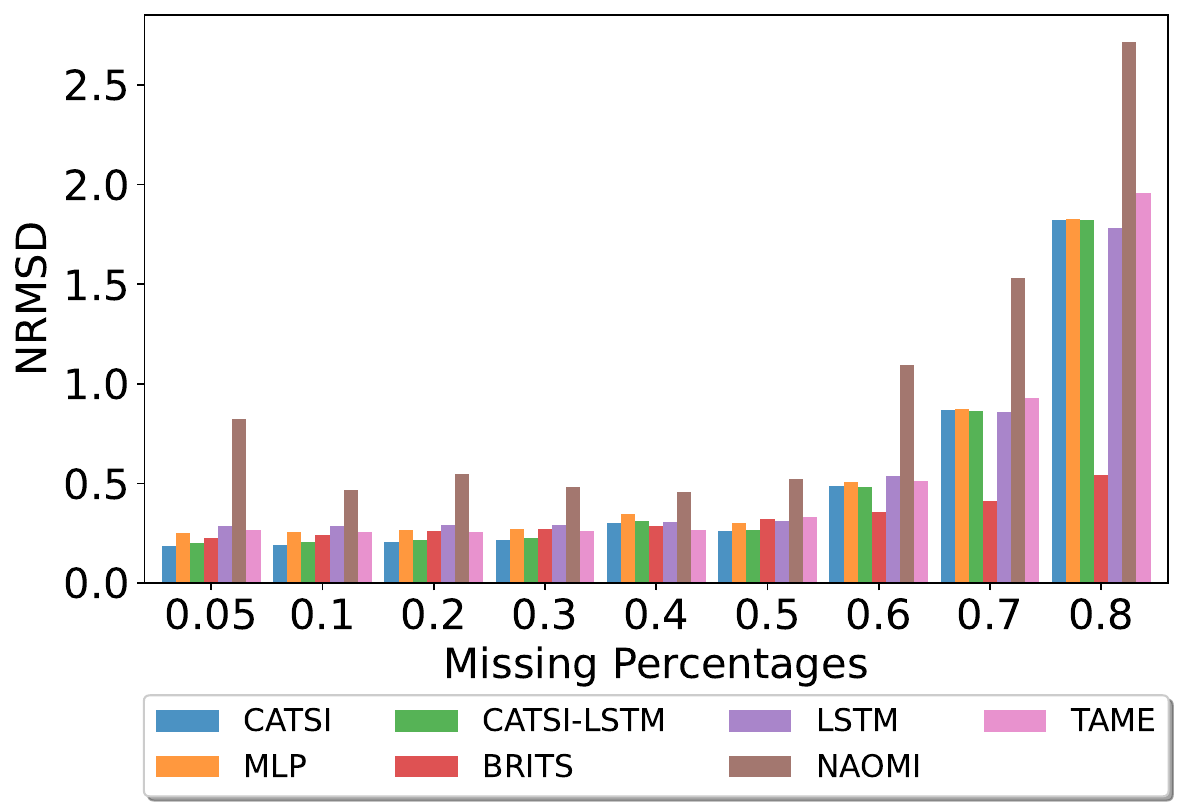} \label{fig:DACMI_MAR}}
\subfigure[Sepsis] { \includegraphics[trim=0.23cm 0.2cm 0.24cm 0.25cm, clip,
width=0.315\textwidth]  {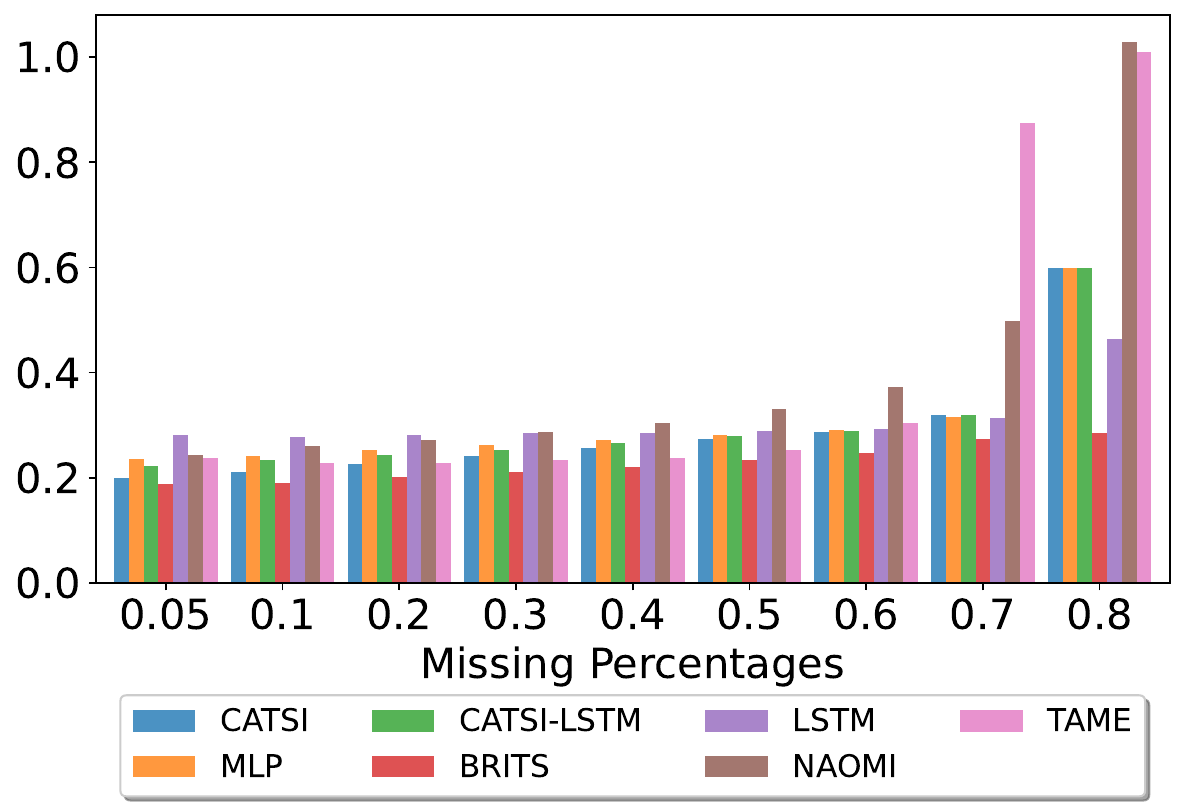} \label{fig:Sepsis_MAR}}
\subfigure[Hypotension] { \includegraphics[trim=0.23cm 0.2cm 0.24cm 0.25cm, clip,
width=0.315\textwidth]  {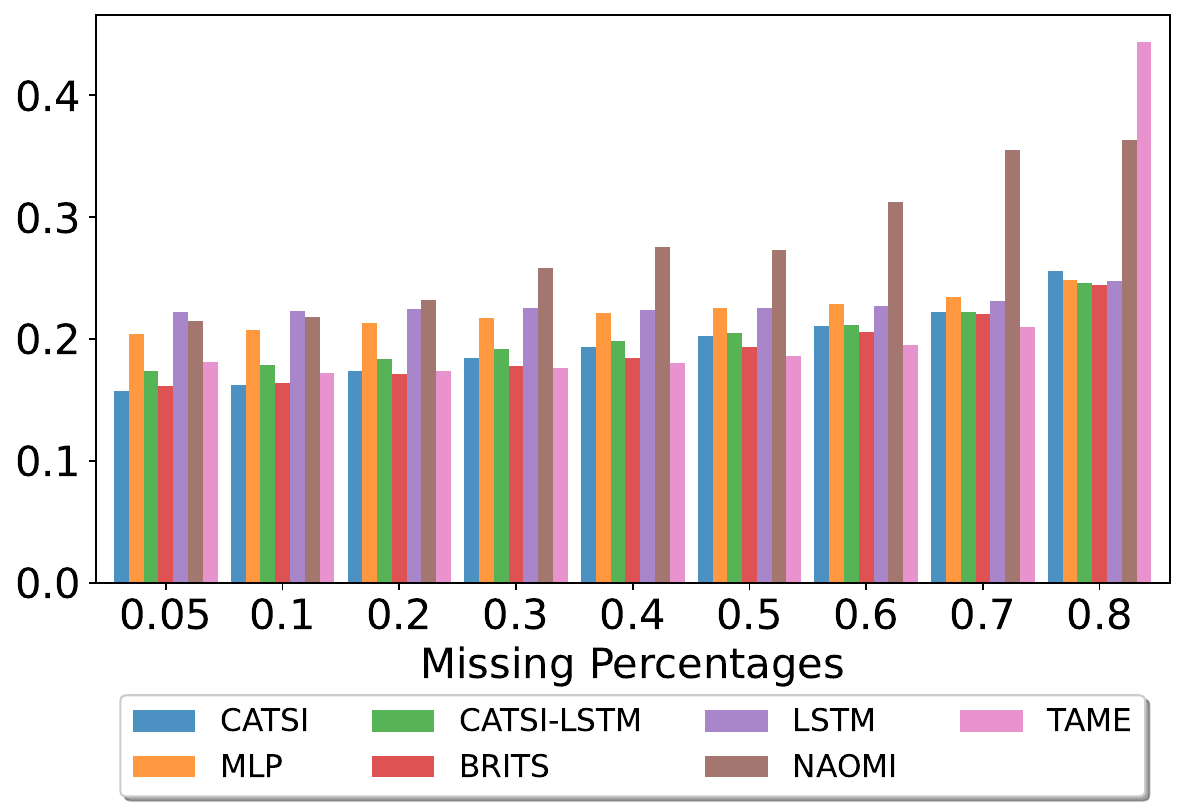} \label{fig:Hyp_MAR}}
\subfigure[IEEEPPG] { \includegraphics[trim=0.23cm 0.2cm 0.24cm 0.25cm, clip,
width=0.315\textwidth]  {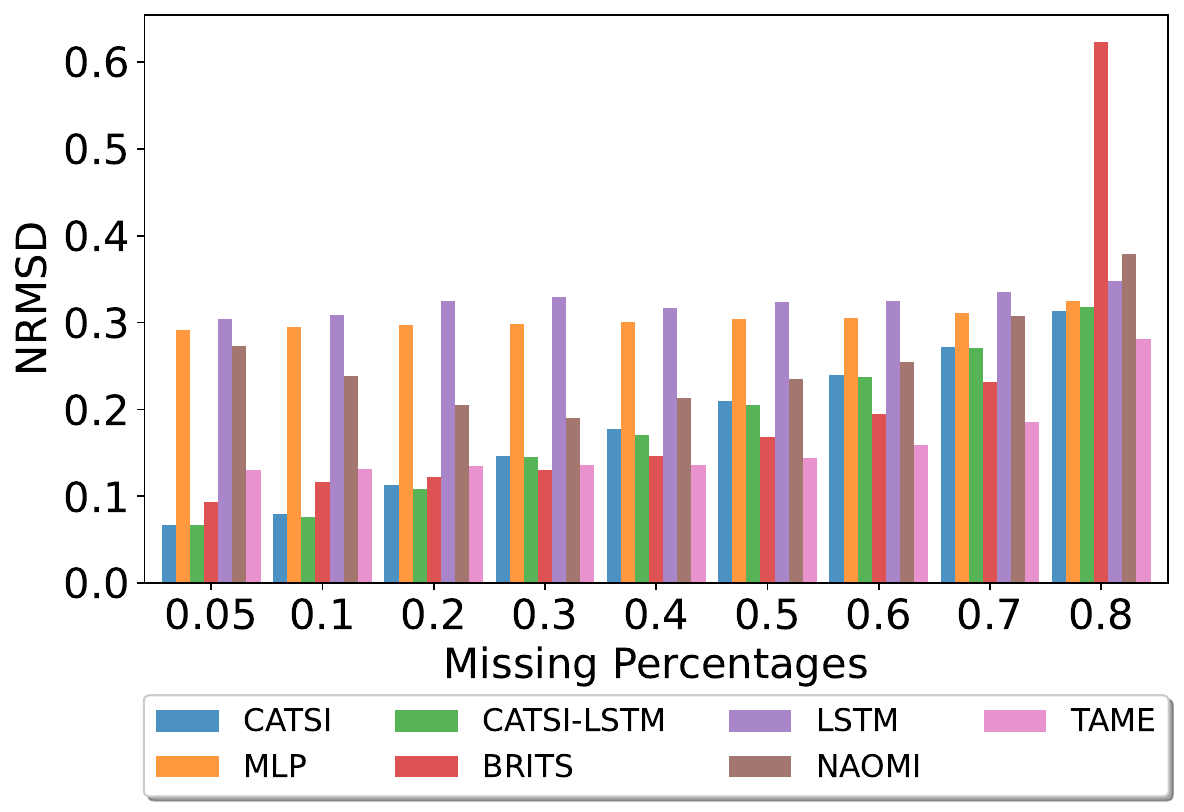} \label{fig:IEEEPPG_MAR}}
\subfigure[Heartbeat] { \includegraphics[trim=0.23cm 0.2cm 0.24cm 0.25cm, clip,
width=0.315\textwidth]  {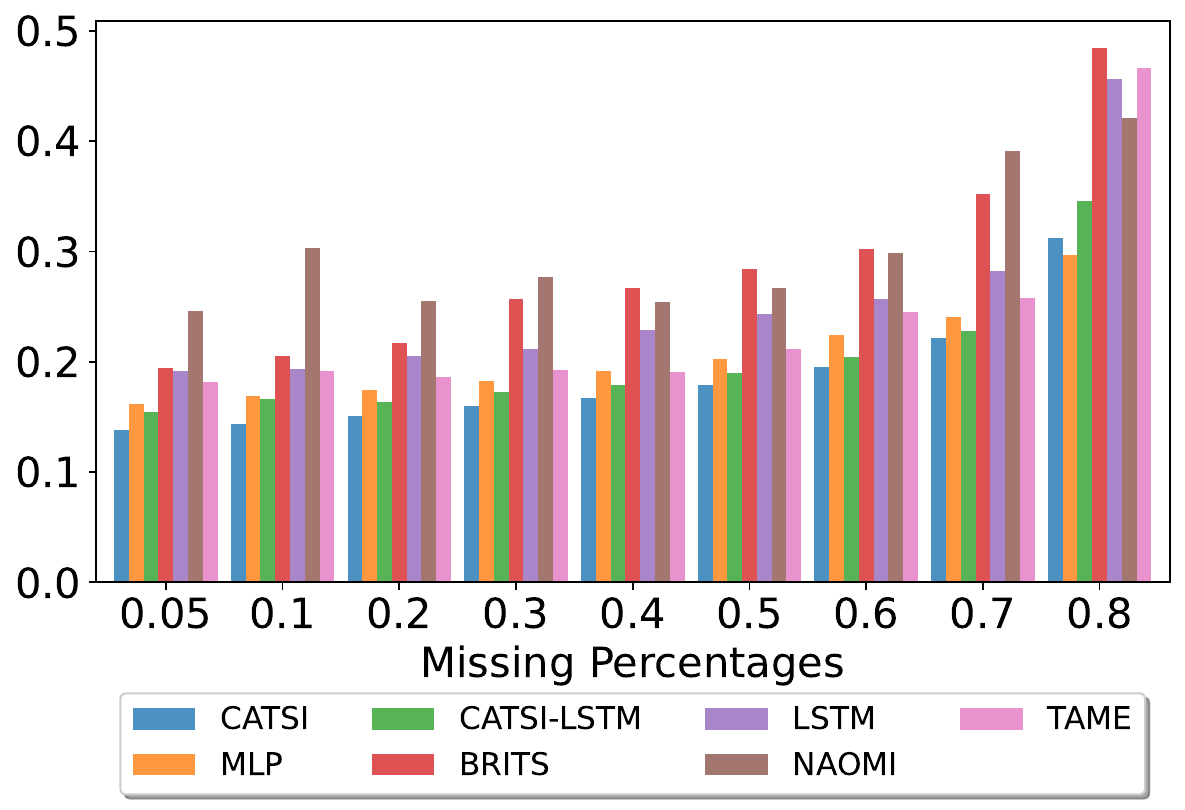} \label{fig:Heart_MAR}}

\caption{Comparing normalized root mean squared deviation (NRMSD) scores of seven deep methods (BRITS, CATSI, CATSI-LSTM, LSTM, CATSI-MLP, NAOMI, TAME) for imputing missing values in time series data. The missing at random (MAR) type is used at varying missing rates.}
\label{fig:MAR}
\end{figure*}


\subsubsection{Hypotension data set}

Table~\ref{tab:cross-time-hypotension} shows that all features except one (MAP) are best imputed using hybrid imputation methods (BRITS and CATSI). The CE-GBT (I and II) method yields the best and second-best imputation accuracy for the MAP and DBP variables, respectively. These two variables have a 0.80 correlation, which explains the superior performance of a cross-sectional imputation method. On the other hand, a moderately correlated (0.71) variable pair (ALT and AST) is better imputed via hybrid and longitudinal imputation methods than cross-sectional. This is because all other variable pairs have a weak correlation (\textless 0.6). Results show that several variables (ATL, AST, urine, and serum creatinine) show the second-best performance using the longitudinal part of the CATSI method, CATSI-LSTM. Overall, integrating cross-sectional and longitudinal imputation approaches in a hybrid framework improves the imputation accuracy for this data set.

\subsubsection{IEEEPPG data set}

None of the cross-sectional imputation methods perform well on the variables of the IEEEPPG data set, as shown in Table~\ref{tab:cross-time-ieeeppg}. All variables are best imputed by the piece-wise time interpolation method, outperforming hybrid methods (CATSI and BRITS). The variables have weak pairwise correlation (highest correlation: 0.31). Therefore, cross-sectional imputation methods are not expected to predict the missing values of other variables accurately. The hybrid CATSI method shows the second-best imputation performance for all variables.

\subsubsection{Heartbeat data set}

The heartbeat data set has 61 variables, which creates a total of 1830 unique variable pairs. Out of 1830 pairs, 315 pairs correlate greater than 0.8, and 69 pairs have a correlation greater than 0.9. The CATSI method is the best for 24 variables and the second best for 20 other variables. The piece-wise time interpolation method provides the second-best imputation performance after the hybrid CATSI method. The piece-wise time interpolation offers the best imputation accuracy for 17 variables and is the second best for 12 other variables.

The CE-GBT-II cross-sectional imputation method performs the best for eight variables and the second best for five other variables. The CE-GBT-I method is the best method for imputing six variables and the second-best for the other five variables. The CATSI-LSTM longitudinal method is the best imputation method for five variables and the second-best for 13 other variables. Out of 69 strongly correlated (\textgreater 0.9) variable pairs (47 unique variables), 14 variables are best imputed by cross-sectional imputation methods (CE-GBT-II: 7; CE-GBT-I: 6; CATSI-MLP: 1).
In general, no single method is superior for imputing missing values of individual variables in time series data. The overall imputation accuracy can be optimized by statistical analysis of the variables and variable-specific selection between cross-sectional, longitudinal, and hybrid imputation methods.


\begin{figure*}[t!]
\centering
\vspace{10pt}
\subfigure[DACMI] { \includegraphics[trim=0.23cm 0.2cm 0.24cm 0.25cm, clip,
width=0.315\textwidth]  {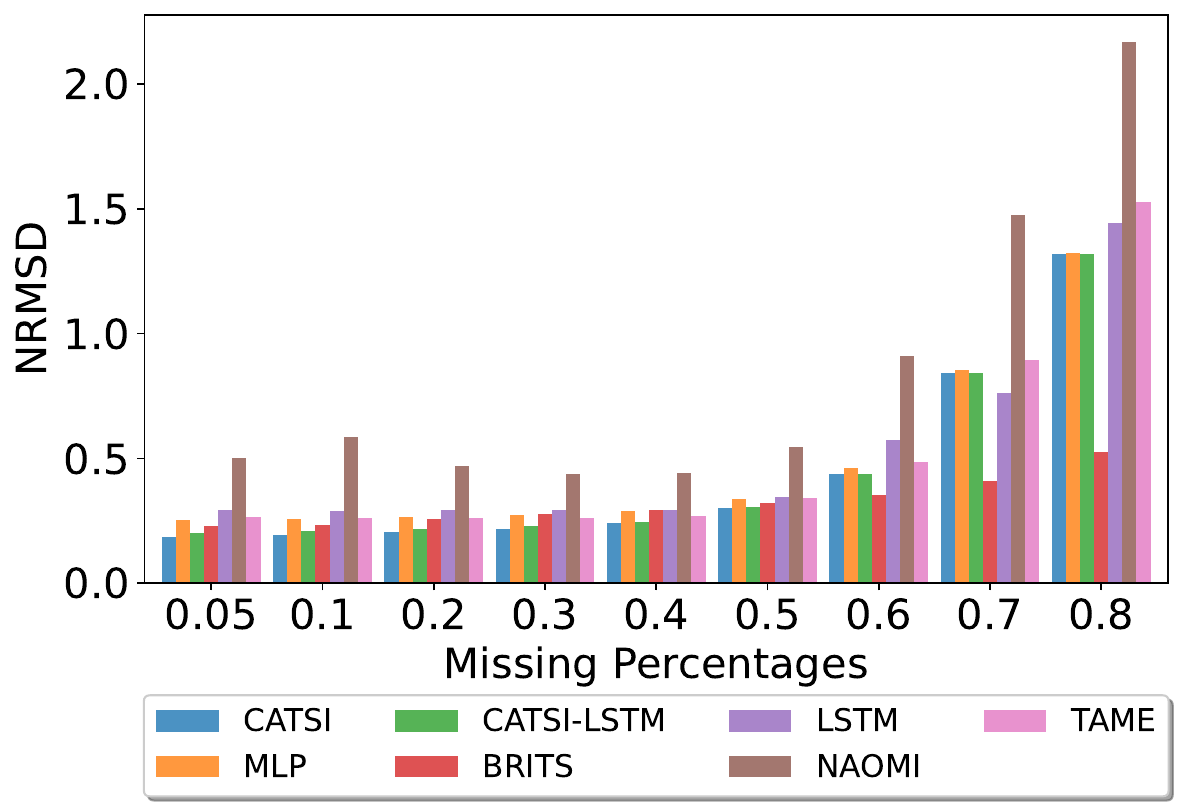} \label{fig:DACMI_MCAR}}
\subfigure[Sepsis] { \includegraphics[trim=0.23cm 0.2cm 0.24cm 0.25cm, clip,
width=0.315\textwidth]  {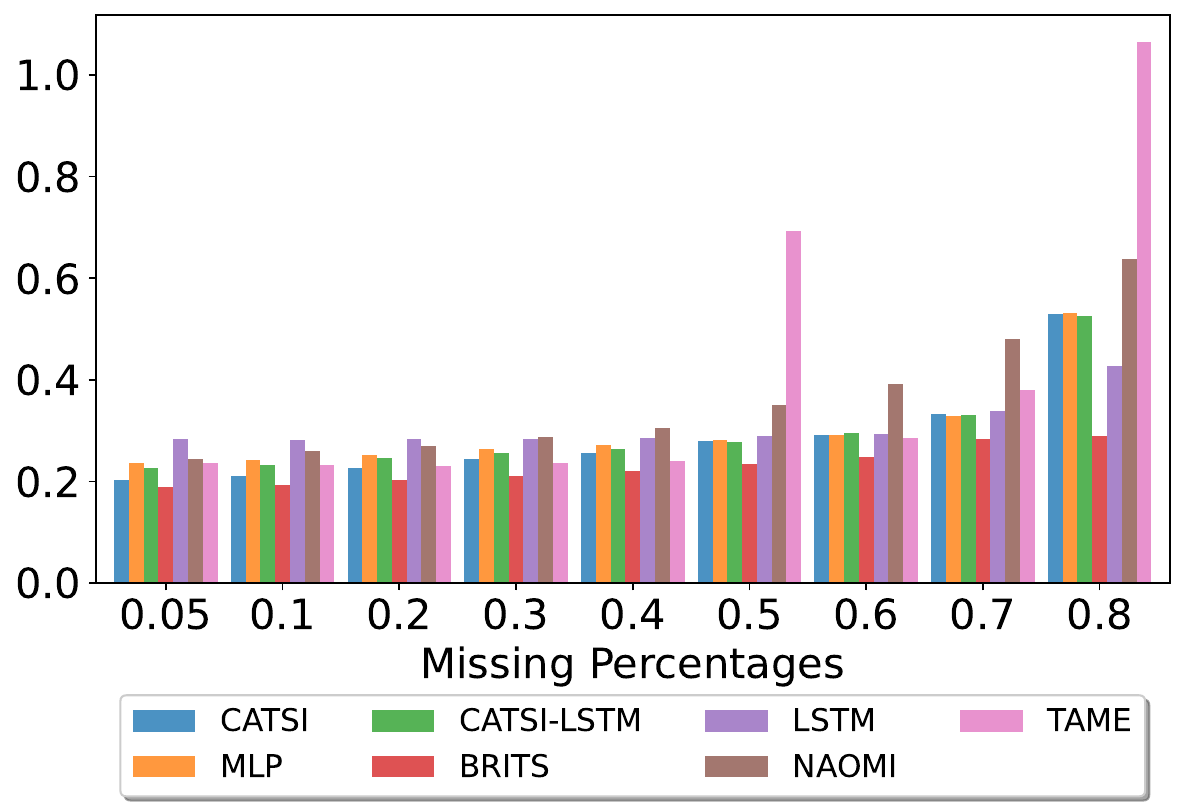} \label{fig:Sepsis_MCAR}}
\subfigure[Hypotension] { \includegraphics[trim=0.23cm 0.2cm 0.24cm 0.25cm, clip,
width=0.315\textwidth]  {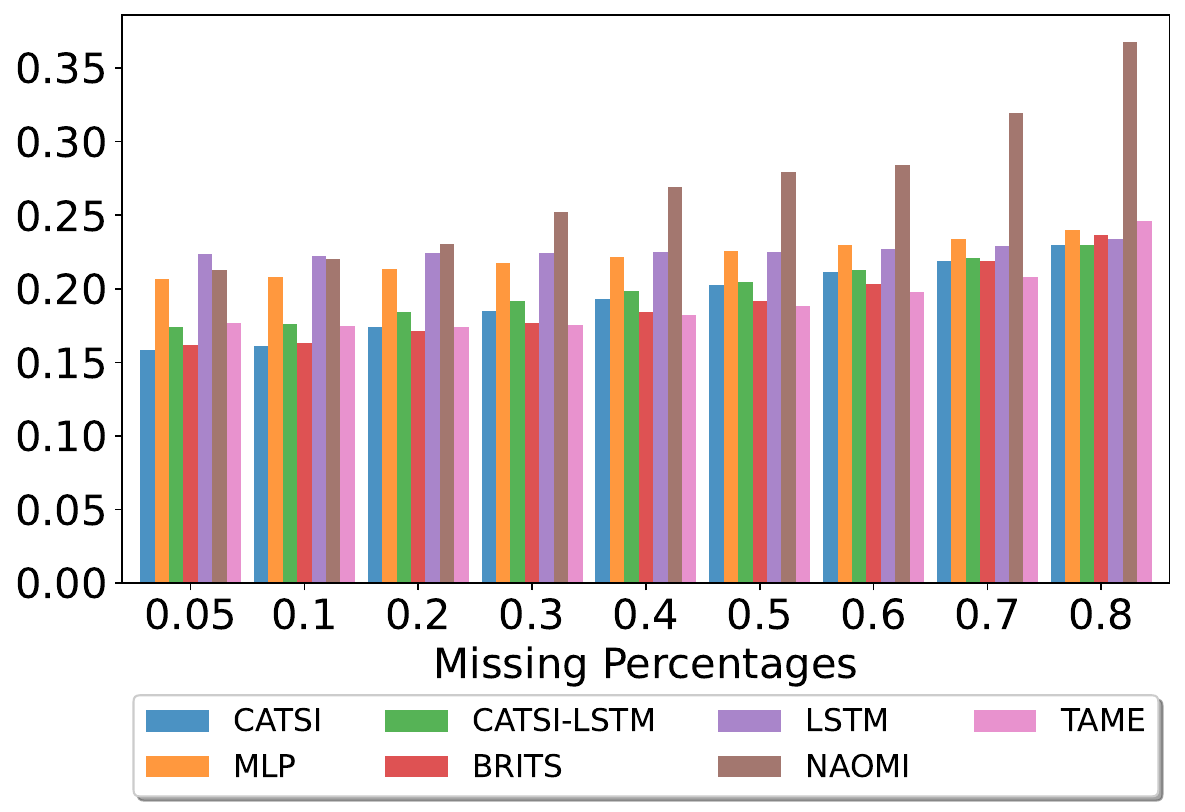} \label{fig:Hyp_MCAR}}
\subfigure[IEEEPPG] { \includegraphics[trim=0.23cm 0.2cm 0.24cm 0.25cm, clip,
width=0.315\textwidth]  {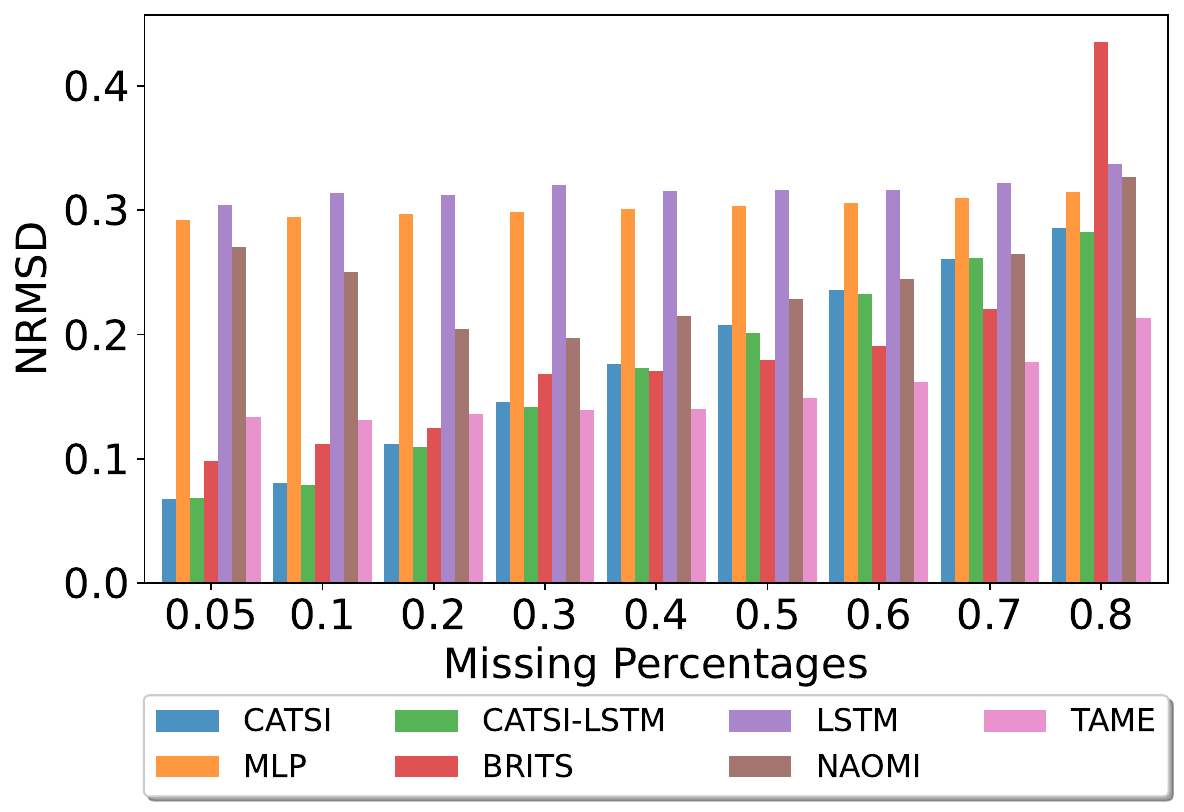} \label{fig:IEEEPPG_MCAR}}
\subfigure[Heartbeat] { \includegraphics[trim=0.23cm 0.2cm 0.24cm 0.25cm, clip,
width=0.315\textwidth]  {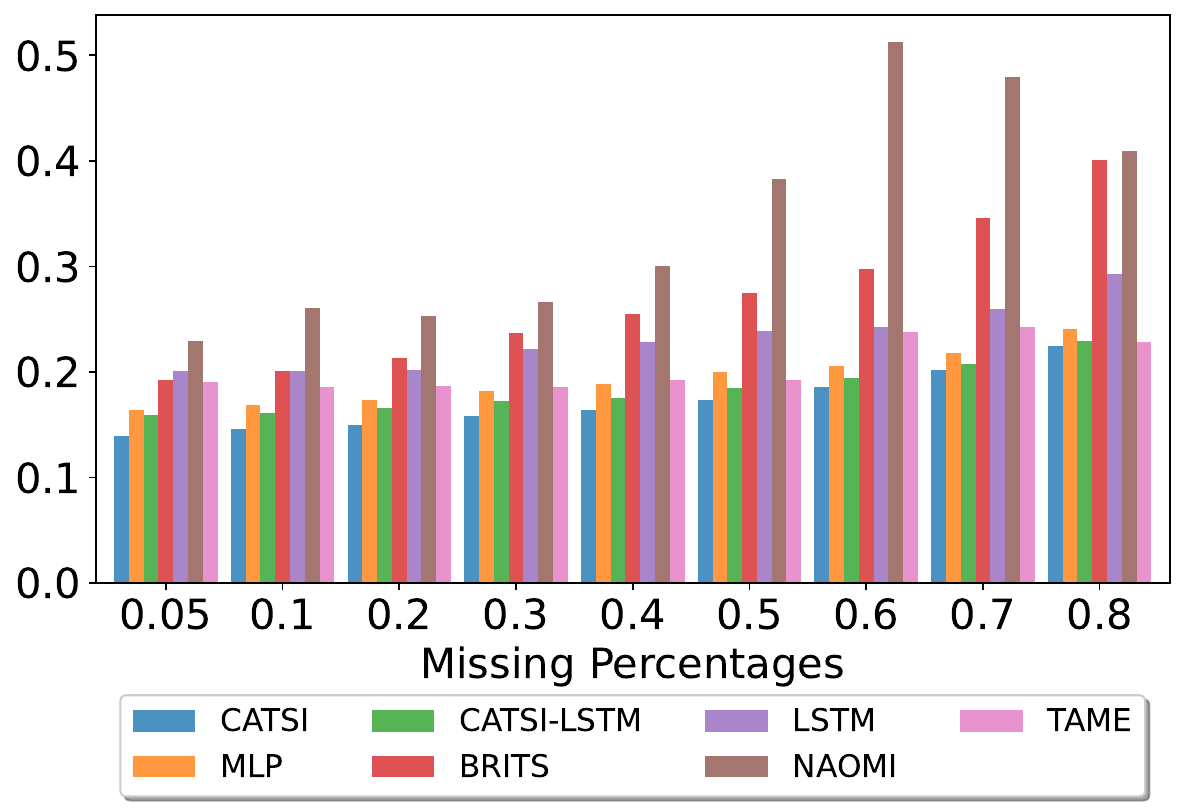} \label{fig:Heart_MCAR}}

\caption{Comparing normalized root mean squared deviation (NRMSD) scores of seven deep methods (BRITS, CATSI, CATSI-LSTM, LSTM, CATSI-MLP, NAOMI, TAME) for imputing missing values in time series data. The missing completely at random (MCAR) type is used at varying missing rates.}
\label{fig:MCAR}
\end{figure*}


\subsection{Effects of missing values rates and types}

When the missing value types are MAR (Figure~\ref{fig:MAR}) and MCAR (Figure~\ref{fig:MCAR}), the CATSI method is the best on the DACMI data set up to 30\% and 50\% missing values, respectively. The BRITS method overtakes CATSI when the missing value rate is more than 50\% for both MAR and MCAR types. In other words, the BRITS method is least susceptible to higher missing value rates for any missing value types. When the missing value type is MNAR, the BRITS method is clearly the superior method for any percentage of missing values on the DACMI data set. The BRITS method appears to be the best for the sepsis data set regardless of the missing value percentages and types. Although the TAME method is known for its high performance, it performs relatively poorly when the missing value percentage is high (Figure~\ref{fig:Sepsis_MAR},~\ref{fig:Sepsis_MCAR}). For the hypotension data set, the CATSI method is the best up to 10\% missing values for MAR and MCAR types, while  
 
\begin{figure*}[t!]
\centering
\vspace{10pt}
\subfigure[DACMI] { \includegraphics[trim=0.23cm 0.2cm 0.24cm 0.25cm, clip,
width=0.315\textwidth]  {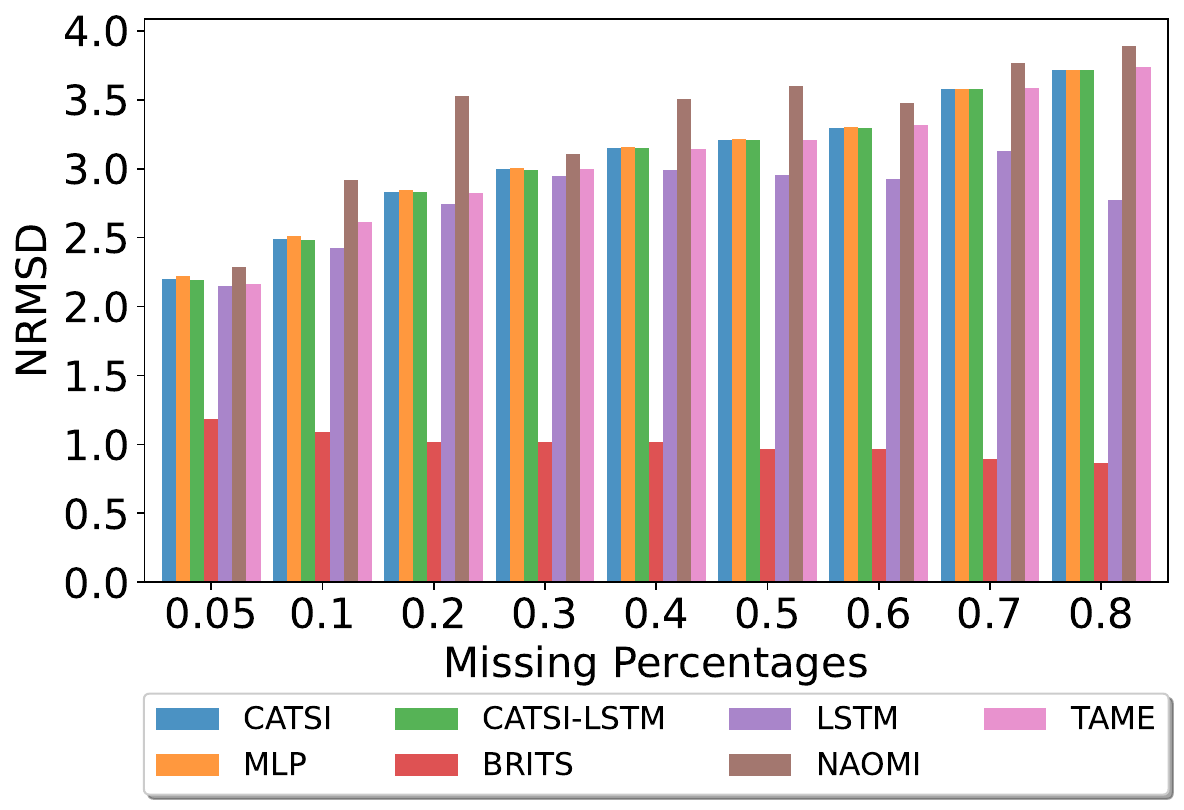} \label{fig:DACMI_MNAR}}
\subfigure[Sepsis] { \includegraphics[trim=0.23cm 0.2cm 0.24cm 0.25cm, clip,
width=0.315\textwidth]  {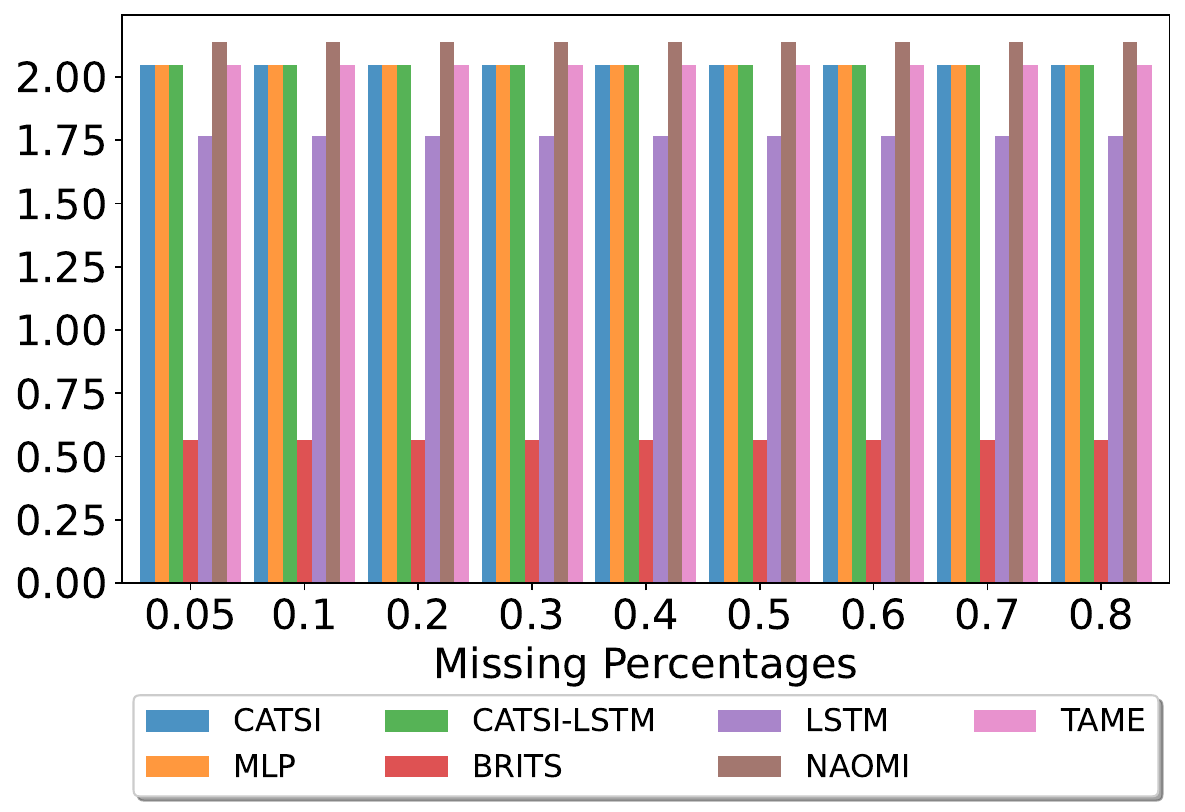} \label{fig:Sepsis_MNAR}}
\subfigure[Hypotension] { \includegraphics[trim=0.23cm 0.2cm 0.24cm 0.25cm, clip,
width=0.315\textwidth]  {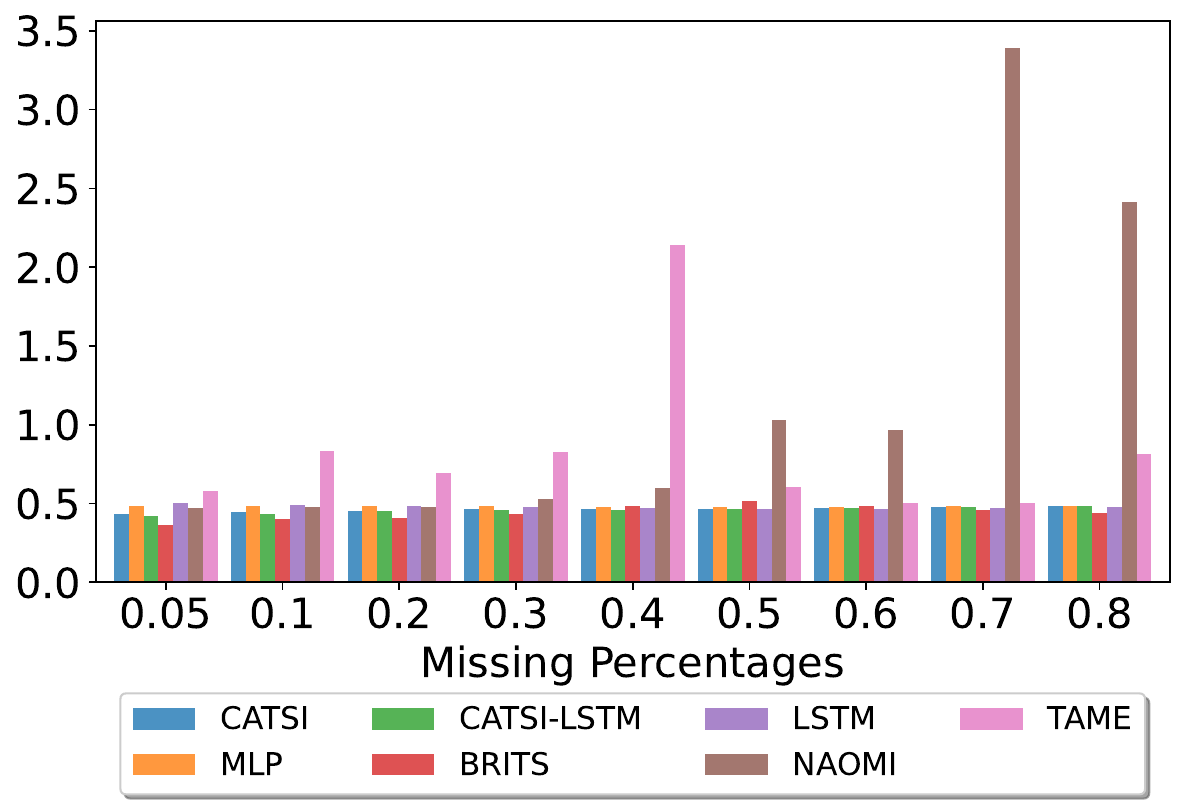} \label{fig:Hyp_MNAR}}
\subfigure[IEEEPPG] { \includegraphics[trim=0.23cm 0.2cm 0.24cm 0.25cm, clip,
width=0.315\textwidth]  {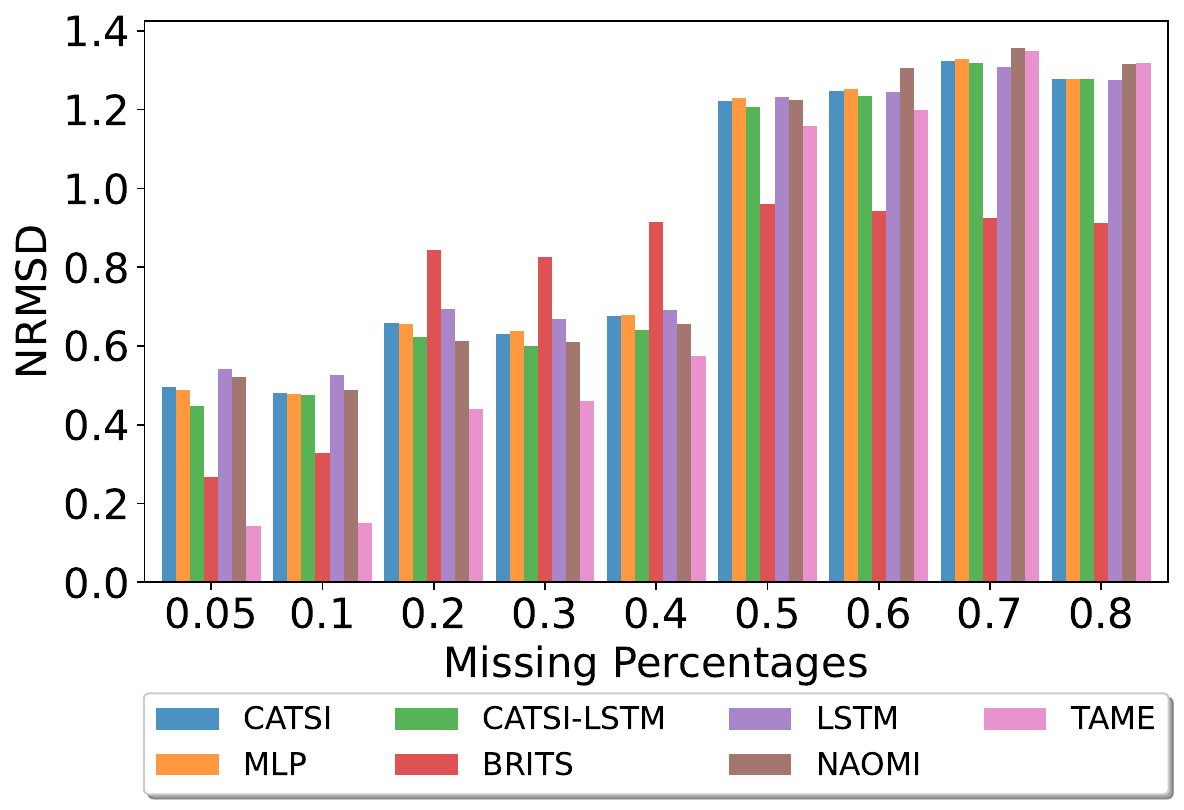} \label{fig:IEEEPPG_MNAR}}
\subfigure[Heartbeat] { \includegraphics[trim=0.23cm 0.2cm 0.24cm 0.25cm, clip,
width=0.315\textwidth]  {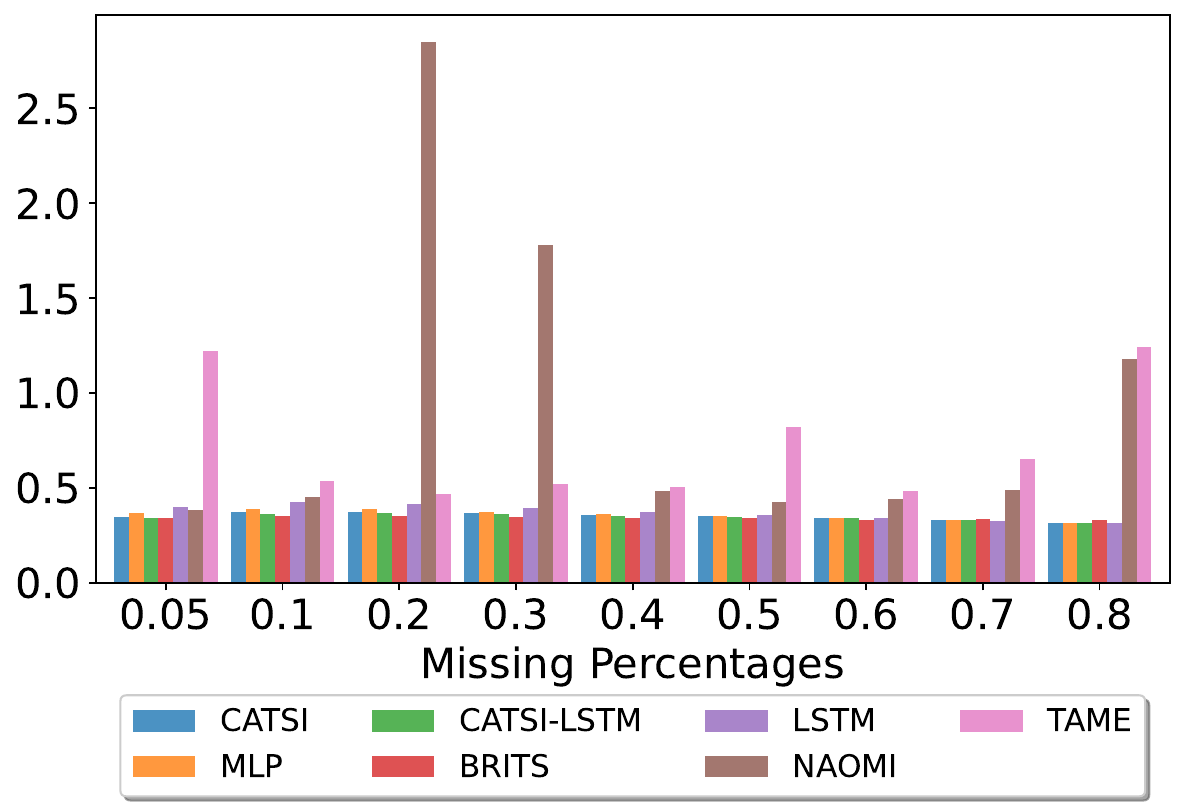} \label{fig:Heart_MNAR}}

\caption{Comparing normalized root mean squared deviation (NRMSD) scores of seven deep methods (BRITS, CATSI, CATSI-LSTM, LSTM, CATSI-MLP, NAOMI, TAME) for imputing missing values in time series data. The missing not-at-random (MNAR) type is used at varying missing rates.}
\label{fig:MNAR}
\end{figure*}

TAME performs the best between 30\% and 70\% missing values. For the MNAR type, the BRITS method is generally superior to all other methods on the hypotension data set, except when the missing rate is between 40\% and 60\% (Figure~\ref{fig:Hyp_MNAR}). The BRITS method performs the best for the MNAR type up to 60\% missing values in the heartbeat data. 

In general, the CATSI method performs the best at all missing rates when the heartbeat data set has missing values of MAR and MCAR types (Figure~\ref{fig:Heart_MAR},~\ref{fig:Heart_MCAR}. The CATSI-LSTM method is the best up to 20\% MAR type missing values in the IEEEPPG data set (Figure~\ref{fig:IEEEPPG_MAR}). The TAME method shows the best performance for 40\% or more MAR type missing values (Figure~\ref{fig:IEEEPPG_MAR}). Similar performance is observed with the MCAR type, where TAME is the best method for 30\% and more missing rates (Figure~\ref{fig:IEEEPPG_MCAR}). The TAME method also performs best up to 40\% MNAR type missing values in the IEEEPPG dataset (Figure~\ref{fig:IEEEPPG_MNAR}). However, the BRITS method outperforms TAME when the missing value rate is 50\% or higher.


\subsection{Classification of imputed data}

\textcolor{black}{The quality of imputed data is evaluated in downstream classification tasks. Three of five data sets (sepsis, hypotension, and heartbeat) have ground truth labels for classification.} The classification label is provided for each time point for the sepsis and hypotension data sets, whereas the heartbeat data set has labels for individual samples. The DACMI and IEEEPPG data sets do not provide classification labels. We perform the following steps to obtain and compare the classification results.

\textcolor{black}{First, we simulate 20\% missing values of MCAR type. The data sets with missing values are then completed using the imputation methods. Second, we randomly sample 80\% of the imputed data to train a classifier. The remaining 20\% data are used for testing and reporting the classification accuracy. This sampling process is repeated ten times to perform statistical analysis of the imputation methods. A gradient-boosting classifier model with a default hyperparameter setting (learning rate: 0.1, number of estimators: 100, maximum tree depth: 3) is trained and tested on imputed data. Third, the quality of imputed data is compared with two baseline scenarios: 1) when the data set is complete with no missing values (CNoMV) and 2) when no imputation method is used to handle the missing values in data~\citep{Razavian2015, Tomavsev2019}, namely the no imputation method (NIM). Tomasev et al. have avoided a method to explicitly impute missing values based on a prior study~\citep{Razavian2015} where the imputation of missing values does not always improve the accuracy of predictive models~\citep{Tomavsev2019}. This observation is based on a multivariate kernel regression-based method in 2015, despite more sophisticated deep learning methods emerging over recent years. Fourth, we choose a binary \emph{readmission} label to classify the sepsis data set. The categorical \emph{vasopressors} label with four class categories is used as the classification target for the hypotension data set. We classify the heartbeat data set using a binary label: the heartbeat as normal versus abnormal.} 

\begin{table}[t]
\caption{\textcolor{black}{The quality of imputed data sets with 20\% missing values of MCAR type is evaluated in downstream classification. The average F1 weighted classification score after sampling ten times is reported. Missing values are initialized using mean values. CNoMV = complete with no missing values; NIM = no imputation.}} 
\resizebox{\textwidth}{!}{\color{black}
\begin{tabular}{ccccccccccccc}
\toprule
\multicolumn{1}{c|}{Data set} & \multicolumn{1}{c|}{BRITS} & \multicolumn{1}{c|}{NAOMI} & \multicolumn{1}{c|}{CATSI} & \multicolumn{1}{c|}{CATSI-LSTM} & \multicolumn{1}{c|}{LSTM} & \multicolumn{1}{c|}{CATSI-MLP} & \multicolumn{1}{c|}{TAME} & \multicolumn{1}{c|}{DETROIT} & \multicolumn{1}{c|}{GP-VAE} & \multicolumn{1}{c|}{MICE} & \multicolumn{1}{c|}{NIM} & CNoMV \\ \midrule
\multicolumn{1}{c|}{Sepsis} & \multicolumn{1}{c|}{\begin{tabular}[c]{@{}c@{}}0.722\\ (0.005)\end{tabular}} & \multicolumn{1}{c|}{\begin{tabular}[c]{@{}c@{}}0.712\\ (0.005)\end{tabular}} & \multicolumn{1}{c|}{\begin{tabular}[c]{@{}c@{}}0.722\\ (0.005)\end{tabular}} & \multicolumn{1}{c|}{\textbf{\begin{tabular}[c]{@{}c@{}}0.724\\ (0.004)\end{tabular}}} & \multicolumn{1}{c|}{\textbf{\begin{tabular}[c]{@{}c@{}}0.724\\ (0.004)\end{tabular}}} & \multicolumn{1}{c|}{\begin{tabular}[c]{@{}c@{}}0.714\\ (0.004)\end{tabular}} & \multicolumn{1}{c|}{\begin{tabular}[c]{@{}c@{}}0.719\\ (0.004)\end{tabular}} & \multicolumn{1}{c|}{\begin{tabular}[c]{@{}c@{}}0.721\\ (0.004)\end{tabular}} & \multicolumn{1}{c|}{\begin{tabular}[c]{@{}c@{}}0.674\\ (0.006)\end{tabular}} & \multicolumn{1}{c|}{\begin{tabular}[c]{@{}c@{}}0.690\\ (0.005)\end{tabular}} & \multicolumn{1}{c|}{\begin{tabular}[c]{@{}c@{}}0.660\\ (0.004)\end{tabular}} & \begin{tabular}[c]{@{}c@{}}0.735\\ (0.005)\end{tabular} \\ \midrule
\multicolumn{1}{c|}{Hypotension} & \multicolumn{1}{c|}{\begin{tabular}[c]{@{}c@{}}0.809\\ (0.002)\end{tabular}} & \multicolumn{1}{c|}{\begin{tabular}[c]{@{}c@{}}0.805\\ (0.003)\end{tabular}} & \multicolumn{1}{c|}{\begin{tabular}[c]{@{}c@{}}0.809\\ (0.002)\end{tabular}} & \multicolumn{1}{c|}{\begin{tabular}[c]{@{}c@{}}0.809\\ (0.003)\end{tabular}} & \multicolumn{1}{c|}{\begin{tabular}[c]{@{}c@{}}0.807\\ (0.002)\end{tabular}} & \multicolumn{1}{c|}{\begin{tabular}[c]{@{}c@{}}0.805\\ (0.002)\end{tabular}} & \multicolumn{1}{c|}{\textbf{\begin{tabular}[c]{@{}c@{}}0.811\\ (0.002)\end{tabular}}} & \multicolumn{1}{c|}{\begin{tabular}[c]{@{}c@{}}0.809\\ (0.001)\end{tabular}} & \multicolumn{1}{c|}{\begin{tabular}[c]{@{}c@{}}0.803\\ (0.002)\end{tabular}} & \multicolumn{1}{c|}{\begin{tabular}[c]{@{}c@{}}0.800\\ (0.002)\end{tabular}} & \multicolumn{1}{c|}{\begin{tabular}[c]{@{}c@{}}0.795\\ (0.002)\end{tabular}} & \begin{tabular}[c]{@{}c@{}}0.811\\ (0.002)\end{tabular} \\ \midrule
\multicolumn{1}{c|}{Heartbeat} & \multicolumn{1}{c|}{\begin{tabular}[c]{@{}c@{}}0.614\\ (0.055)\end{tabular}} & \multicolumn{1}{c|}{\begin{tabular}[c]{@{}c@{}}0.614\\ (0.042)\end{tabular}} & \multicolumn{1}{c|}{\begin{tabular}[c]{@{}c@{}}0.651\\ (0.067)\end{tabular}} & \multicolumn{1}{c|}{\textbf{\begin{tabular}[c]{@{}c@{}}0.659\\ (0.061)\end{tabular}}} & \multicolumn{1}{c|}{\begin{tabular}[c]{@{}c@{}}0.638\\ (0.054)\end{tabular}} & \multicolumn{1}{c|}{\begin{tabular}[c]{@{}c@{}}0.635\\ (0.078)\end{tabular}} & \multicolumn{1}{c|}{\begin{tabular}[c]{@{}c@{}}0.632\\ (0.053)\end{tabular}} & \multicolumn{1}{c|}{\begin{tabular}[c]{@{}c@{}}0.637\\ (0.029)\end{tabular}} & \multicolumn{1}{c|}{\begin{tabular}[c]{@{}c@{}}0.604\\ (0.056)\end{tabular}} & \multicolumn{1}{c|}{\begin{tabular}[c]{@{}c@{}}0.586\\ (0.064)\end{tabular}} & \multicolumn{1}{c|}{\begin{tabular}[c]{@{}c@{}}0.603\\ (0.035)\end{tabular}} & \begin{tabular}[c]{@{}c@{}}0.603\\ (0.049)\end{tabular} \\ \midrule
\end{tabular}
}
\label{tab:classification}
\end{table}

\subsubsection{Classification accuracy}Table~\ref{tab:classification} shows the \textcolor{black}{average} F1 weighted classification score for each imputation method for the mean initialization setting on three data sets. The sepsis data set imputed by the \textcolor{black}{CATSI-LSTM and LSTM methods shows the best classification performance (weighted F1 score: 0.724 (0.004))}. \textcolor{black}{The worst classification accuracy is obtained by the no imputation method (NIM) (weighted F1 score: 0.660 (0.004)). Intuitively, the complete data set with no missing values (CNoMV) yields the highest weighted F1 score of 0.735 (0.005).}

For the hypotension data set, the classification accuracy on imputed test data is similar across different imputation methods. \textcolor{black}{The TAME method yields the best classification performance (weighted F1 score: 0.811 (0.002)). Complete data set with no missing values (CNoMV) achieves the same accuracy as the data imputed by the TAME method. Data with missing values but without using any explicit imputation method shows the worst classification accuracy for hypotension data set(weighted F1 score: 0.795 (0.002).}
 
\textcolor{black}{The heartbeat data set shows the best classification performance with CATSI-LSTM imputed data  (weighted F1 score: 0.659 (0.061)), which is better than the accuracy of the complete data set with no missing values (weighted F1 score: 0.603 (0.049)). All other deep imputation methods have similarly shown better  classification accuracy than using complete signals without missing values. This observation remains consistent in multiple experiments. Unlike EHR data (Sepsis or Hypotension), the heartbeat data set is a collection of sensor signals possibly with redundant frequency and noise components. It is possible that introducing missing values in such signal and then imputing can filter out noise components in the signal, improving overall signal quality. The MICE method shows the worst classification accuracy for the heartbeat data set (weighted F1 score: 0.586 (0.064)). }

\begin{table}[t]
\caption{\textcolor{black}{Statistical comparison between imputation methods using the sepsis data set. Methods in the row are read as statistically better (B) or worse (W) than those in the columns. NS = not significant, CNoMV = complete with no missing values, NIM = no imputation.}} 
\resizebox{\textwidth}{!}{\color{black}
\begin{tabular}{c|cccccccccccc}
\toprule
Method & BRITS & NAOMI & CATSI & CATSI-LSTM & LSTM & CATSI-MLP & TAME & DETROIT & GP-VAE & MICE & NIM & CNoMV \\ \midrule
BRITS & - & p\textless{}0.05, B & NS & NS & NS & p\textless{}0.05, B & p\textless{}0.05, B & NS & p\textless{}0.05, B & p\textless{}0.05, B & p\textless{}0.05, B & p\textless{}0.05, W \\
NAOMI &  & - & p\textless{}0.05, W & p\textless{}0.05, W & p\textless{}0.05, W & NS & p\textless{}0.05, W & p\textless{}0.05, W & p\textless{}0.05, B & p\textless{}0.05, B & p\textless{}0.05, B & p\textless{}0.05, W \\
CATSI &  &  & - & NS & NS & p\textless{}0.05, B & NS & NS & p\textless{}0.05, B & p\textless{}0.05, B & p\textless{}0.05, B & p\textless{}0.05, W \\
CATSI-LSTM &  &  &  & - & NS & p\textless{}0.05, B & p\textless{}0.05, B & NS & p\textless{}0.05, B & p\textless{}0.05, B & p\textless{}0.05, B & p\textless{}0.05, W \\
LSTM &  &  &  &  & - & p\textless{}0.05, B & p\textless{}0.05, B & NS & p\textless{}0.05, B & p\textless{}0.05, B & p\textless{}0.05, B & p\textless{}0.05, W \\
CATSI-MLP &  &  &  &  &  & - & p\textless{}0.05, W & p\textless{}0.05, W & p\textless{}0.05, B & p\textless{}0.05, B & p\textless{}0.05, B & p\textless{}0.05, W \\
TAME &  &  &  &  &  &  & - & NS & p\textless{}0.05, B & p\textless{}0.05, B & p\textless{}0.05, B & p\textless{}0.05, W \\
DETROIT &  &  &  &  &  &  &  & - & p\textless{}0.05, B & p\textless{}0.05, B & p\textless{}0.05, B & p\textless{}0.05, W \\
GP-VAE &  &  &  &  &  &  &  &  & - & p\textless{}0.05, W & p\textless{}0.05, B & p\textless{}0.05, W \\
MICE &  &  &  &  &  &  &  &  &  & - & p\textless{}0.05, B & p\textless{}0.05, W \\
NIM &  &  &  &  &  &  &  &  &  &  & - & p\textless{}0.05, W \\
\bottomrule
\end{tabular}
}
\label{tab:wilcoxon-sepsis}
\end{table}

\begin{table}[t]
\caption{\textcolor{black}{Statistical comparison between imputation methods using the hypotension data set. Methods in the row are read as statistically better (B) or worse (W) than those in the columns. NS = not significant, CNoMV = complete with no missing values, NIM = no imputation.} }
\resizebox{\textwidth}{!}{\color{black}
\begin{tabular}{c|cccccccccccc}
\toprule
Method & BRITS & NAOMI & CATSI & CATSI-LSTM & LSTM & CATSI-MLP & TAME & DETROIT & GP-VAE & MICE & NIM & CNoMV \\ \midrule
BRITS & - & p\textless{}0.05, B & NS & NS & p\textless{}0.05, B & p\textless{}0.05, B & NS & NS & p\textless{}0.05, B & p\textless{}0.05, B & p\textless{}0.05, B & p\textless{}0.05, W \\
NAOMI &  & - & p\textless{}0.05, W & p\textless{}0.05, W & NS & NS & p\textless{}0.05, W & p\textless{}0.05, W & NS & p\textless{}0.05, B & p\textless{}0.05, B & p\textless{}0.05, W \\
CATSI &  &  & - & NS & NS & p\textless{}0.05, B & NS & NS & p\textless{}0.05, B & p\textless{}0.05, B & p\textless{}0.05, B & NS \\
CATSI-LSTM &  &  &  & - & NS & p\textless{}0.05, B & NS & NS & p\textless{}0.05, B & p\textless{}0.05, B & p\textless{}0.05, B & p\textless{}0.05, W \\
LSTM &  &  &  &  & - & NS & p\textless{}0.05, W & p\textless{}0.05, W & p\textless{}0.05, B & p\textless{}0.05, B & p\textless{}0.05, B & p\textless{}0.05, W \\
CATSI-MLP &  &  &  &  &  & - & p\textless{}0.05, W & p\textless{}0.05, W & p\textless{}0.05, B & p\textless{}0.05, B & p\textless{}0.05, B & p\textless{}0.05, W \\
TAME &  &  &  &  &  &  & - & NS & p\textless{}0.05, B & p\textless{}0.05, B & p\textless{}0.05, B & NS \\
DETROIT &  &  &  &  &  &  &  & - & p\textless{}0.05, B & p\textless{}0.05, B & p\textless{}0.05, B & NS \\
GP-VAE &  &  &  &  &  &  &  &  & - & p\textless{}0.05, B & p\textless{}0.05, B & p\textless{}0.05, W \\
MICE &  &  &  &  &  &  &  &  &  & - & p\textless{}0.05, B & p\textless{}0.05, W \\
NIM &  &  &  &  &  &  &  &  &  &  & - & p\textless{}0.05, W \\
\bottomrule
\end{tabular}
}
\label{tab:wilcoxon-hypotension}
\end{table}


\subsubsection{Statistical analysis of classification accuracy}
\textcolor{black}{Train and test data are randomly sampled ten times from imputed data to perform Wilcoxon signed-rank test on classification accuracies.} \textcolor{black}{The sepsis data set (Table~\ref{tab:wilcoxon-sepsis}) shows that the classification of data without missing values (CNoMV) is statistically better than that of imputed data by any methods. However, all imputation methods yield statistically better classification than the case with missing values without imputation (NIH). All imputation methods are statistically better than the MICE method, except the GP-VAE, which is statistically worse than MICE. The CATSI method is better than CATSI-MLP, NAOMI, GP-VAE, and MICE. However, the CATSI method shows no statistically different performance compared to CATSI-LSTM, LSTM, TAME, and DETROIT methods.}

\textcolor{black}{The hypotension data set (Table~\ref{tab:wilcoxon-hypotension}) shows that the classification accuracy on data without missing values is statistically similar to those imputed by the CATSI, TAME, and DETROIT methods.} \textcolor{black}{All imputation methods are statistically better than the no imputation method (NIM). Similarly, all imputation methods are statistically better than MICE imputation. All methods except NAOMI are statistically better than GP-VAE. The classification of data imputed by NAOMI, LSTM, and CATSI-MLP methods is statistically worse than the ones imputed by TAME and DETROIT methods. The BRITS method has significantly better results than NAOMI, LSTM, and CATSI-MLP methods. Furthermore, the hypotension data set imputed by the NAOMI method produces significantly worse results than CATSI and CATSI-LSTM methods. Lastly, BRITS, CATSI, and CATSI-LSTM has significantly better results than CATSI-MLP.}

\textcolor{black}{In general, the classification accuracies are not statistically different when comparing two imputation methods on the heartbeat data set. However, CATSI-LSTM yields statistically better accuracy than the NAOMI. An LSTM-based imputation is statistically superior to the GP-VAE, MICE, NIM, and CNoMV methods. The DETROIT method is statistically superior to the MICE and NIM methods.}

\section{Discussion}

This paper surveys and benchmarks state-of-the-art deep learning methods for imputing missing values in time-series health data. The findings of this work can be summarized as follows. First, no single imputation method is the best for all types of health data. For EHR data, hybrid imputation methods such as CATSI or BRITS may be preferred because these methods generally perform the best on the DACMI, sepsis, and hypotension data sets. For low-dimensional and weakly correlated signal data like the IEEEPPG data set, longitudinal methods such as CATSI-LSTM or simple piece-wise time interpolation may be preferred. Longitudinal imputation methods are also promising for strongly correlated time series signals (heartbeat data set). Second, initializing missing values with a simple piece-wise time interpolation has proved to be the best option in four of five data sets (Table~\ref{tab:misssing-value-init-3}). Specifically, piece-wise time initialization shows superiority in initializing missing values on time series signal data. Third, the CATSI imputation method is superior for time series data with a length of 15 and more (Table~\ref{tab:time-length}). For time series data with fewer than 10 time points, a cross-sectional imputation method may be more effective. Fourth, no single method can achieve the best imputation accuracy for all variables within a specific data set. The variables with high cross-correlation with other variables are generally better imputed by a cross-sectional imputation method than a hybrid or longitudinal method. However, hybrid imputation methods generally show the best accuracy for imputing individual variables. Furthermore, individual variable imputation accuracy can depend on the data domain. For example, longitudinal imputation methods (\textcolor{black}{e.g., CATSI-LSTM, LSTM}) usually perform the best on variables of time series signal data (Table~\ref{tab:cross-time-ieeeppg}).



Fifth, the imputation performance on randomly missing values (MAR, MCAR) and  non-randomly missing value types (MNAR) is substantially different. The imputation error is substantially higher on the MNAR type (Figure~\ref{fig:MNAR}) than on other types (MAR, MCAR) (Figures~\ref{fig:MAR},~\ref{fig:MCAR}). This aligns with the previous observation that data with not-at-random missing values are more challenging to impute correctly~\citep{Samad2022}. This finding is important because missing values in EHR data \textcolor{black}{do not appear randomly as the MCAR type}. Therefore, state-of-the-art missing value imputation methods benchmarked on data with MCAR type may not demonstrate equally superior accuracy on real-world data.  

Sixth, the BRITS method generally yields the best imputation accuracy for the MNAR missing data type across varying missing rates. For the MCAR and MAR data types, BRITS also performs well on high missing rates ($>$50\%) on several data sets (DACMI, sepsis, hypotension). The TAME method performs better at high missing rates ($>$50\%) than at lower. In contrast, the CATSI method performs much better at lower missing rates ($<$50\%) than higher ones. Therefore, the superiority of state-of-the-art time series data imputation methods, often evaluated on low missing rates, may not work equally well on real-world data with high missing rates.

\textcolor{black}{Finally, the effect of imputation on downstream patient classification tasks is mixed. For EHR data sets (sepsis, hypotension), three deep imputation methods (CATSI, TAME, and DETROIT) generally yield statistically better classification results than other baseline methods, including MICE and no imputation strategies. Similar statistical significance is not observed for heartbeat sensor signal data. However, sensor signal data consistently reveal an unusual observation that imputed data sets can yield better classification accuracy than those without missing values. We attribute this observation to the redundant frequency and noise components in sensor data that may be filtered out due to the removal and imputation of signal values. Therefore, the data removal and imputation process can help improve the quality of sensor signals for downstream classification tasks.}

 \begin{table}[t]
\caption{\textcolor{black}{Execution times (in minutes) for training and testing of 13 imputation methods on five time series health data sets (DACMI, Sepsis, Hypotension, IEEEPPG, Heartbeat).}}
\resizebox{\textwidth}{!}{\color{black} 
\begin{tabular}{c|c|c|c|c|c|c|c|c|c|c|c|c|c}
\toprule
Dataset & BRITS & CATSI & CATSI-LSTM & LSTM & MLP & NAOMI & TAME & DETROIT & GP-VAE & MICE & CE-GBT-I & CE-GBT-II & PW \\ \midrule
DACMI & 190.5 & 79.7 & 73.3 & 54.6 & 17.6 & 28.9 & 36.4 & 12.5 & 66.4 & 1.3 & 15.3 & 207.9 & 6.1 \\
Sepsis & 22.6 & 8.6 & 8.8 & 6.9 & 3.8 & 4.6 & 9.1 & 9.6 & 1.8 & 0.4 & 48.7 & 681.2 & 3.7 \\
Hypotension & 64.5 & 25.9 & 24.8 & 19.2 & 6.9 & 10.8 & 13.8 & 9.5 & 0.9 & 0.6 & 15.9 & 291.8 & 4.8 \\
IEEEPPG & 51.9 & 21.4 & 19.7 & 15.1 & 5.1 & 10.1 & 8.2 & 14.8 & 3.2 & 0.3 & 2.1 & 36.6 & 2.1 \\
Heartbeat & 9.1 & 4.2 & 3.4 & 3.0 & 1.7 & 4.8 & 377.1 & 4.2 & 2.4 & 0.5 & 65.3 & 1321.5 & 3.3 \\ \bottomrule
\end{tabular}
}
\label{tab:time-complexity}
\end{table}

\subsection{\textcolor{black}{Recommendations for health scientists}}
\textcolor{black}{Health scientists use conventional methods to handle missing values in healthcare data, including no imputation, complete case analysis by removing samples with missing values, and imputing values with mean or median. MICE-based imputation methods are arguably the most popular approach in biostatistics and appear in more recent clinical studies on retrospective healthcare data~\citep{Samad2018}. In contrast, health science often avoids superior deep learning methods due to high computational costs, availability of such methods in easy-to-use software packages (e.g., SAS) or languages (e.g., R), and poor interpretability. Table~\ref{tab:time-complexity} compares the time required for 13 different imputation strategies on the five data sets. Indeed, deep learning methods are more time-consuming than MICE  and other conventional baselines. However, missing value imputation is a crucial part of preparing real-world data, which happens only once to ensure data quality and the reliability of data-driven outcomes. Because high-performance computing on the cloud and local servers is a reality, ensuring top-quality data should be a high priority instead of adopting an arbitrary method for imputing missing values.
\begin{itemize}
    \item Health data most commonly appear as multivariate time series, whereas MICE methods model relationships between variables, ignoring essential temporal relationships in time series. The most basic imputation strategy should involve piece-wise time interpolation followed by MICE-based cross-sectional imputation. 
    \item Hybrid methods that learn cross-sectional and longitudinal relationships in data are preferred to ensure the highest quality data. We recommend the CATSI or BRITS methods because the TAME and DETROIT methods require ground truth values for model training, which may not be available in practice.  
    \item Hybrid deep imputation methods generally show imputation and classification accuracy statistically superior to conventional baselines. However, the improvement may not be statistically significant in all scenarios. Despite statistical insignificance, a 1\% relative improvement in classification accuracy infers correctly prognosticating 100 additional patients in 10,000 samples, which is valuable in clinical practice. Therefore, statistical tests should not always dissuade researchers from adopting advanced imputation methods to ensure the highest data quality. 
\item Missing values are less of a concern in health data acquired using digital sensors than in structured EHR data tables. There may not be any missing values or need for imputation in sensor signal data. However, removing and imputing signal values have consistently improved data quality in downstream classification due to a potential noise-filtering effect. Therefore, sensor signals used in digital health may be subject to removal and imputation processes to augment the data quality.
\item We recommend using high-performance computing, including cloud-based or local GPU servers, to reap the best fruits of deep learning effectively.
\item The implementation of open-source deep imputation methods in Python remains a challenge for researchers trained in R programming and statistical software packages. The productive transfer of knowledge and tools is an open problem and would entail interdisciplinary efforts and research support.   
    \end{itemize}}

\section{Conclusions}

This study presents one of the first surveys to benchmark state-of-the-art deep learning methods for imputing missing values in time series data. Our six data-centric experiments compare the effectiveness of imputation methods across different data types, variable statistics, time series length, missing value rates and types, and \textcolor{black}{in downstream classification tasks}. \textcolor{black}{The experimental results have identified several important insights into method selection for imputing time series data and shared recommendations for health scientists.  Therefore, we emphasize the importance of considering data-centric factors over an arbitrary selection of an imputation strategy to ensure data quality and the reliability of data-driven outcomes. In the future, the interactions between the six experimental conditions may be an important direction of study for optimizing the data-centric selection of imputation methods.}




\section*{Acknowledgements}

Research reported in this publication was supported by the National Library Of Medicine of the National Institutes of Health under Award Number R15LM013569. The content is solely the responsibility of the authors and does not necessarily represent the official views of the National Institutes of Health.

\bibliographystyle{elsarticle-num}
\bibliography{bibliography}

\newpage
\section*{\textcolor{black}{Appendix A. Effects of missing value initialization methods}}\label{appendix:a}

\renewcommand\thetable{A.1} 
\begin{table}[h]
\caption{\textcolor{black}{Effects of missing value initialization methods on individual imputation models and data sets using average normalized root mean squared deviation scores (NRMSD) and average rank across all variables.}} 
\resizebox{\textwidth}{!}{\begin{tabular}{cc|cc|cc|cc|cc|cc|cc}
\toprule
\multicolumn{2}{c|}{Dataset} & \multicolumn{2}{c|}{DACMI} & \multicolumn{2}{c|}{Sepsis} & \multicolumn{2}{c|}{Hypotension} & \multicolumn{2}{c|}{IEEEPPG} & \multicolumn{2}{c|}{Heartbeat} & \multicolumn{2}{c}{\begin{tabular}[c]{@{}c@{}}Avg. perf.\\ \textcolor{black}{across} all\\ data sets\end{tabular}} \\ \midrule
\multicolumn{1}{c|}{Method} & Initialization & NRMSD & \begin{tabular}[c]{@{}c@{}}Avg.\\ rank\end{tabular} & NRMSD & \begin{tabular}[c]{@{}c@{}}Avg.\\ rank\end{tabular} & NRMSD & \begin{tabular}[c]{@{}c@{}}Avg.\\ rank\end{tabular} & NRMSD & \begin{tabular}[c]{@{}c@{}}Avg.\\ rank\end{tabular} & NRMSD & \begin{tabular}[c]{@{}c@{}}Avg\\ rank\end{tabular} & NRMSD & Rank \\ \midrule
\multicolumn{1}{c|}{\multirow{4}{*}{BRITS}} & Delta & 0.266 & 12.2 & 0.190 & 5.9 & 0.163 & 7.0 & 0.094 & 11.6 & 0.199 & 16.4 & 0.182 & 12 \\
\multicolumn{1}{c|}{} & Mean & 0.267 & 11.5 & 0.190 & 5.8 & 0.163 & 7.4 & 0.094 & 12.2 & 0.195 & 16.2 & 0.182 & 9 \\
\multicolumn{1}{c|}{} & Median & 0.267 & 12.6 & 0.190 & 6.0 & 0.163 & 7.6 & 0.094 & 11.8 & 0.195 & 15.9 & 0.182 & 10 \\
\multicolumn{1}{c|}{} & Piece-wise & 0.270 & 12.7 & \textbf{0.190} & \textbf{5.3} & 0.163 & 8.4 & 0.094 & 12.0 & 0.194 & 15.4 & 0.182 & 11 \\ \midrule
\multicolumn{1}{c|}{\multirow{4}{*}{CATSI}} & Delta & \textbf{0.206} & \textbf{3.7} & 0.202 & 6.1 & 0.158 & 5.4 & 0.064 & 5.6 & 0.140 & 4.3 & 0.154 & 2 \\
\multicolumn{1}{c|}{} & Mean & 0.207 & 5.1 & 0.202 & 6.3 & 0.158 & 5.1 & 0.065 & 7.2 & 0.140 & 4.3 & 0.154 & 3 \\
\multicolumn{1}{c|}{} & Median & 0.207 & 5.9 & 0.203 & 6.9 & 0.158 & 5.1 & 0.065 & 6.6 & 0.141 & 4.9 & 0.155 & 4 \\
\multicolumn{1}{c|}{} & Piece-wise & 0.207 & 5.9 & 0.203 & 6.5 & \textbf{0.158} & \textbf{4.1} & 0.057 & 2.8 & \textbf{0.139} & \textbf{3.9} & \textbf{0.153} & \textbf{1} \\ \midrule
\multicolumn{1}{c|}{\multirow{4}{*}{CATSI-LSTM}} & Delta & 0.223 & 8.6 & 0.226 & 13.5 & 0.174 & 11.6 & 0.063 & 4.8 & 0.157 & 10.5 & 0.169 & 6 \\
\multicolumn{1}{c|}{} & Mean & 0.223 & 8.5 & 0.227 & 13.7 & 0.173 & 10.0 & 0.068 & 10.2 & 0.159 & 11.7 & 0.170 & 8 \\
\multicolumn{1}{c|}{} & Median & 0.224 & 9.2 & 0.227 & 13.7 & 0.174 & 11.3 & 0.064 & 6.6 & 0.157 & 10.2 & 0.169 & 7 \\
\multicolumn{1}{c|}{} & Piece-wise & 0.223 & 9.0 & 0.226 & 13.0 & 0.173 & 10.9 & \textbf{0.053} & \textbf{1.2} & 0.156 & 9.8 & 0.166 & 5 \\ \midrule
\multicolumn{1}{c|}{\multirow{4}{*}{LSTM}} & Delta & 0.298 & 17.6 & 0.292 & 24.9 & 0.224 & 21.7 & 0.289 & 18.2 & 0.191 & 16.6 & 0.259 & 18 \\
\multicolumn{1}{c|}{} & Mean & 0.310 & 19.4 & 0.282 & 22.8 & 0.224 & 21.1 & 0.317 & 24.4 & 0.196 & 17.9 & 0.266 & 19 \\
\multicolumn{1}{c|}{} & Median & 0.323 & 21.7 & 0.288 & 24.3 & 0.228 & 24.1 & 0.319 & 25.0 & 0.194 & 17.1 & 0.270 & 20 \\
\multicolumn{1}{c|}{} & Piece-wise & 0.241 & 12.5 & 0.363 & 26.7 & 0.239 & 24.9 & 0.080 & 10.4 & 0.173 & 12.2 & 0.219 & 13 \\ \midrule
\multicolumn{1}{c|}{\multirow{4}{*}{CATSI-MLP}} & Delta & 0.268 & 15.1 & 0.233 & 15.4 & 0.204 & 19.0 & 0.295 & 20.6 & 0.164 & 12.6 & 0.233 & 14 \\
\multicolumn{1}{c|}{} & Mean & 0.268 & 15.5 & 0.234 & 15.7 & 0.204 & 18.1 & 0.295 & 19.8 & 0.164 & 12.4 & 0.233 & 15 \\
\multicolumn{1}{c|}{} & Median & 0.268 & 15.6 & 0.234 & 15.8 & 0.204 & 18.1 & 0.295 & 20.8 & 0.164 & 12.6 & 0.233 & 16 \\
\multicolumn{1}{c|}{} & Piece-wise & 0.265 & 13.6 & 0.238 & 17.3 & 0.205 & 19.4 & 0.294 & 19.2 & 0.164 & 12.7 & 0.233 & 17 \\ \midrule
\multicolumn{1}{c|}{\multirow{4}{*}{NAOMI}} & Delta & 0.688 & 25.2 & 0.242 & 17.4 & 0.218 & 18.8 & 0.259 & 15.0 & 0.245 & 23.1 & 0.330 & 23 \\
\multicolumn{1}{c|}{} & Mean & 0.617 & 25.4 & 0.242 & 17.5 & 0.218 & 18.7 & 0.273 & 16.2 & 0.247 & 23.4 & 0.319 & 22 \\
\multicolumn{1}{c|}{} & Median & 0.437 & 22.9 & 0.240 & 17.2 & 0.220 & 19.4 & 0.285 & 18.2 & 0.251 & 23.7 & 0.287 & 21 \\
\multicolumn{1}{c|}{} & Piece-wise & 0.824 & 26.0 & 0.270 & 21.9 & 0.228 & 23.1 & 0.054 & 2.0 & 0.288 & 26.1 & 0.333 & 24 \\ \midrule
\multicolumn{1}{c|}{\multirow{4}{*}{\textcolor{black}{MICE}}} & Delta & 0.730 & 22.2 & 0.252 & 17.2 & 0.337 & 22.1 & 0.901 & 26.9 & 0.332 & 24.4 & 0.510 & 26 \\
\multicolumn{1}{c|}{} & Mean & 0.733 & 23.8 & 0.252 & 17.2 & 0.337 & 22.1 & 0.901 & 26.6 & 0.332 & 24.5 & 0.511 & 28 \\
\multicolumn{1}{c|}{} & Median & 0.730 & 23.3 & 0.252 & 16.9 & 0.337 & 22.1 & 0.901 & 27.4 & 0.332 & 25.0 & 0.510 & 27 \\
\multicolumn{1}{c|}{} & Piece-wise & 0.730 & 21.9 & 0.252 & 16.9 & 0.337 & 21.0 & 0.901 & 25.1 & 0.332 & 24.1 & 0.510 & 25 \\ \bottomrule
\end{tabular}}
\label{tab:misssing-value-init-2}
\end{table}

\newpage
\section*{\textcolor{black}{Appendix B. Correlation between variables versus imputation accuracy (NRMSD scores) to demonstrate the selection of cross-sectional (MLP) and longitudinal (CATSI-LSTM) imputation methods for individual variables.}}\label{appendix:b}

\renewcommand\thefigure{B.1} 

\begin{figure*}[h!]
\centering
\vspace{10pt}

\subfigure[DACMI - HCT variable] { \includegraphics[trim=0.23cm 0.2cm 0.24cm 0.25cm, clip,
width=0.48\textwidth]  {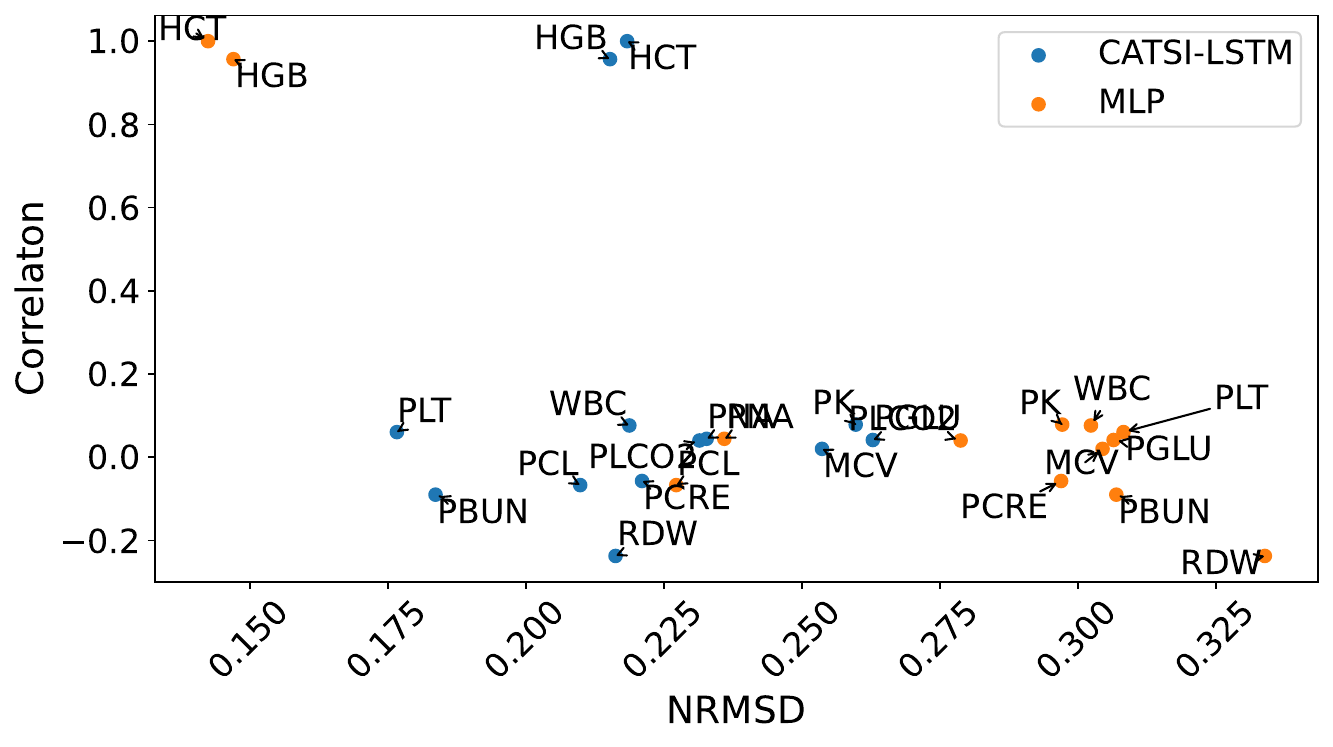} \label{app:dac-hct}}

\subfigure[Hypotension - DBP variable] { \includegraphics[trim=0.23cm 0.2cm 0.24cm 0.25cm, clip,
width=0.48\textwidth]  {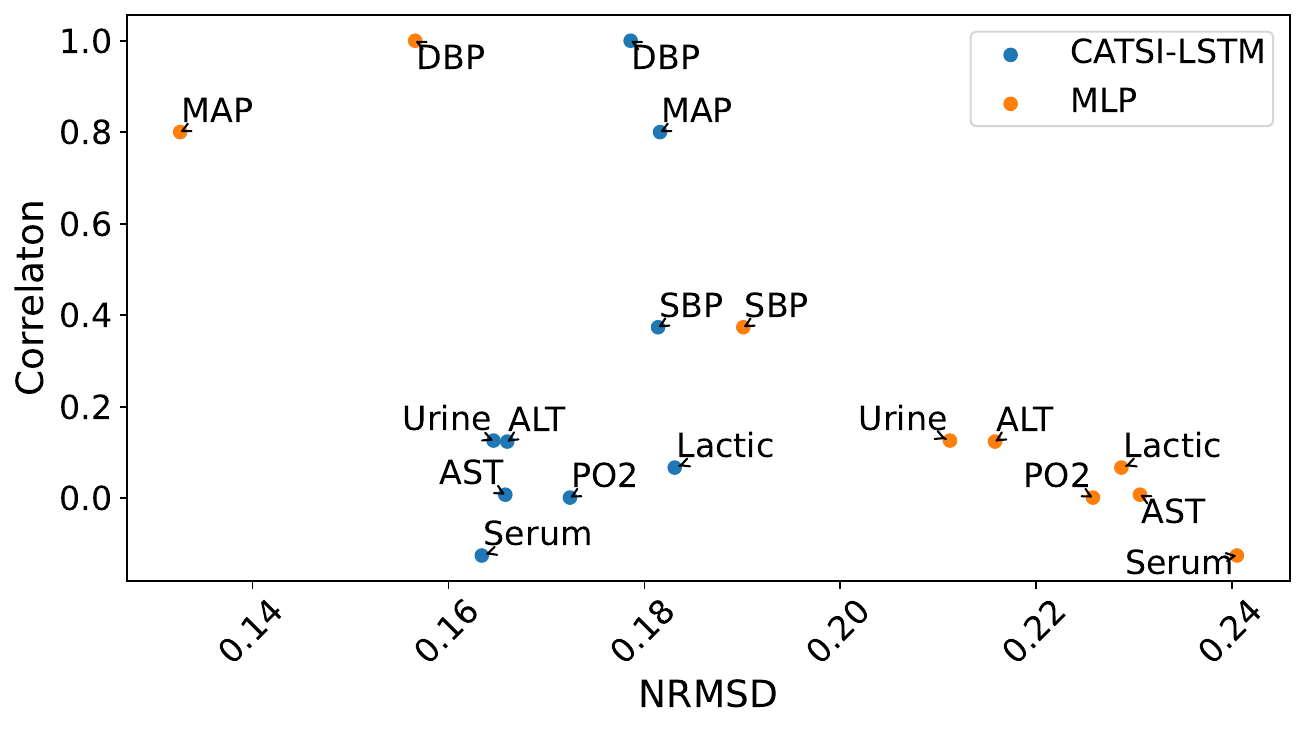} \label{app:hyp-dbp}}
\caption{\textcolor{black}{(a) Correlation between HCT and other variables in the DACMI data set versus the NRMSD scores for individual variables, (b) the same plot for the DBP variable in the Hypotension data set. }}
\label{fig:dac}
\end{figure*}

\end{document}